\documentclass[letterpaper]{article} 
\usepackage{aaai25}  
\usepackage{times}  
\usepackage{helvet}  
\usepackage{courier}  
\usepackage[hyphens]{url}  
\usepackage{graphicx} 
\usepackage{natbib}  
\usepackage{caption} 
\usepackage{algorithm}
\usepackage{algorithmic}
\usepackage{subcaption}
\usepackage{multirow}
\usepackage{graphicx}
\usepackage{float}
\usepackage{booktabs}
\usepackage{pifont}
\usepackage[table,xcdraw]{xcolor}
\usepackage{makecell}
\usepackage{siunitx} 
\usepackage{newfloat}
\usepackage{listings}
\DeclareCaptionStyle{ruled}{labelfont=normalfont,labelsep=colon,strut=off} 
\floatstyle{ruled}
\newfloat{listing}{tb}{lst}{}
\floatname{listing}{Listing}
\title{FriendsQA: A New Large-Scale Deep Video Understanding Dataset with\\ Fine-grained Topic Categorization for Story Videos}
\author{
    Zhengqian Wu\textsuperscript{\rm 1,\rm 3,\rm 4}\equalcontrib,
    Ruizhe Li\textsuperscript{\rm 1,\rm 3,\rm 4}\equalcontrib, 
    Zijun Xu\textsuperscript{\rm 2,\rm 3,\rm 4},\\
    Zhongyuan Wang\textsuperscript{\rm 1,\rm 2,\rm 3,\rm 4}, 
    Chunxia Xiao\textsuperscript{\rm 1,\rm 2,\rm 3,\rm 4},
    Chao Liang\textsuperscript{\rm 1,\rm 2,\rm 3,\rm 4}\thanks{Chao Liang is the corresponding author.}
}
\affiliations{
    \textsuperscript{\rm 1}School of Computer Science, Wuhan University, China\\
    \textsuperscript{\rm 2}School of Cyber Science and Engineering, Wuhan University, China \\
    \textsuperscript{\rm 3}National Engineering Research Center for Multimedia Software, Wuhan University, China \\
    \textsuperscript{\rm 4}Hubei Key Laboratory of Multimedia and Network Communication Engineering, Wuhan University, China \\
    \{2023202110068, 2020300004016, 2019300003083, cxxiao, cliang\}@whu.edu.cn, wzy\_hope@163.com
}
\usepackage{bibentry}

\begin{document}

\maketitle

\begin{abstract}
Video question answering (VideoQA) aims to answer natural language questions according to the given videos. Although existing models perform well in the factoid VideoQA task, they still face challenges in deep video understanding (DVU) task, which focuses on story videos. Compared to factoid videos, the most significant feature of story videos is storylines, which are composed of complex interactions and long-range evolvement of core story topics including characters, actions and locations. Understanding these topics requires models to possess DVU capability. However, existing DVU datasets rarely organize questions according to these story topics, making them difficult to comprehensively assess VideoQA models' DVU capability of complex storylines. Additionally, the question quantity and video length of these dataset are limited by high labor costs of handcrafted dataset building method. In this paper, we devise a large language model based multi-agent collaboration framework, StoryMind, to automatically generate a new large-scale DVU dataset. The dataset, FriendsQA, derived from the renowned sitcom \textit{Friends} with an average episode length of 1,358 seconds, contains 44.6K questions evenly distributed across 14 fine-grained topics. Finally, We conduct comprehensive experiments on 10 state-of-the-art VideoQA models using the FriendsQA dataset.
\end{abstract}

%
\begin{links}
    \link{Code and Datasets}{https://github.com/nercms-mmap/FriendsQA}
\end{links}

\section{Introduction}
Video question answering (VideoQA) aims to answer natural language questions based on given videos, supporting advanced applications like video grounding \cite{timechat} and video chatbots \cite{chat-univi}. Early researches on factoid VideoQA focus on questions about actions or objects in videos \cite{activitynet-qa}. With the popularity of video language model (VLM) and multimodal large language model (MLLM) technologies, significant progress has been made in recent years \cite{sevila,videoqasurvey}.

\begin{figure}[t]
\centering
\includegraphics[width=\linewidth]{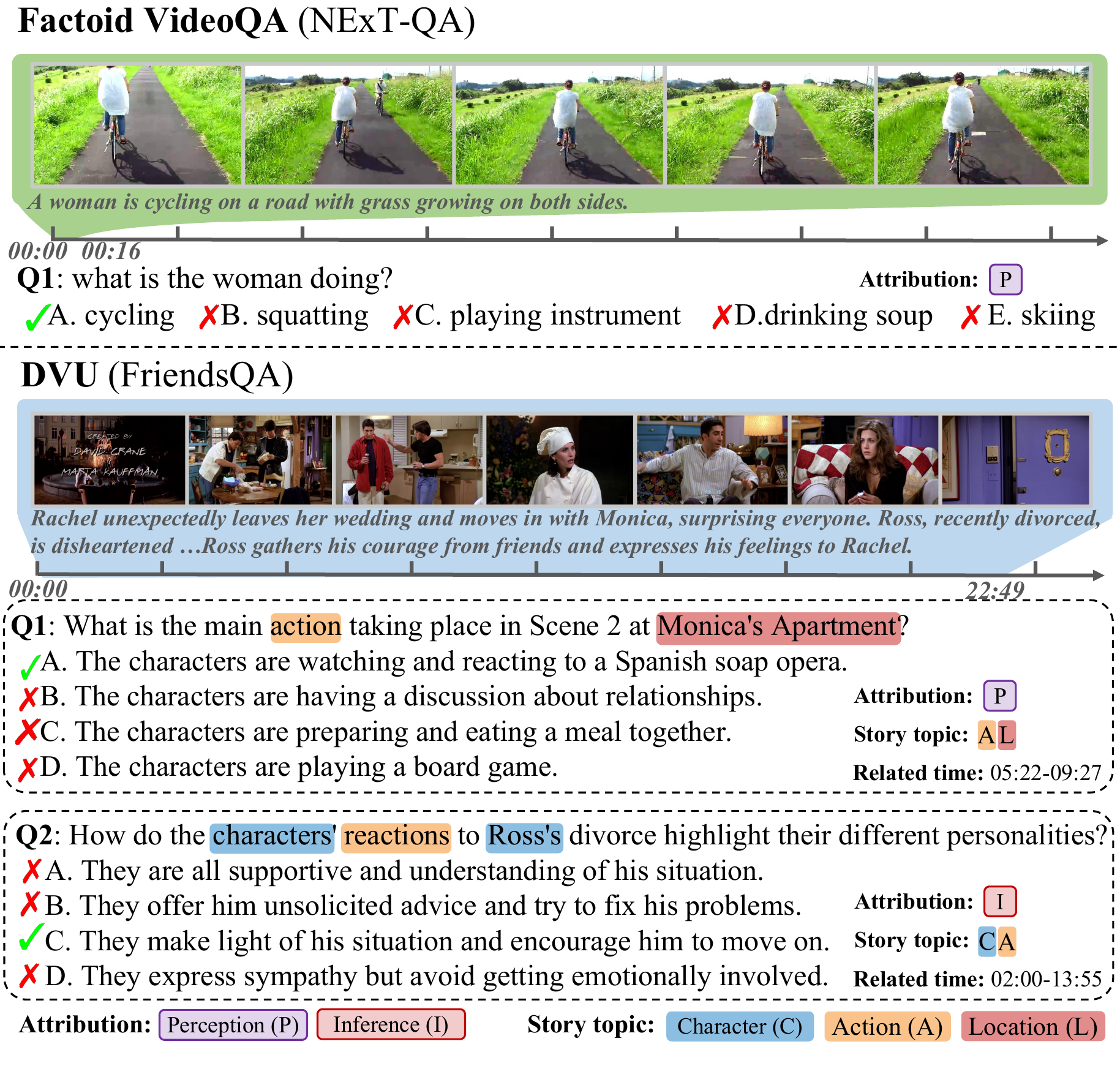} 
\caption{Comparisons of factoid VideoQA and DVU.}
\label{fig:example}
\end{figure}

\begin{figure}[t]
\centering
\includegraphics[width=\linewidth]{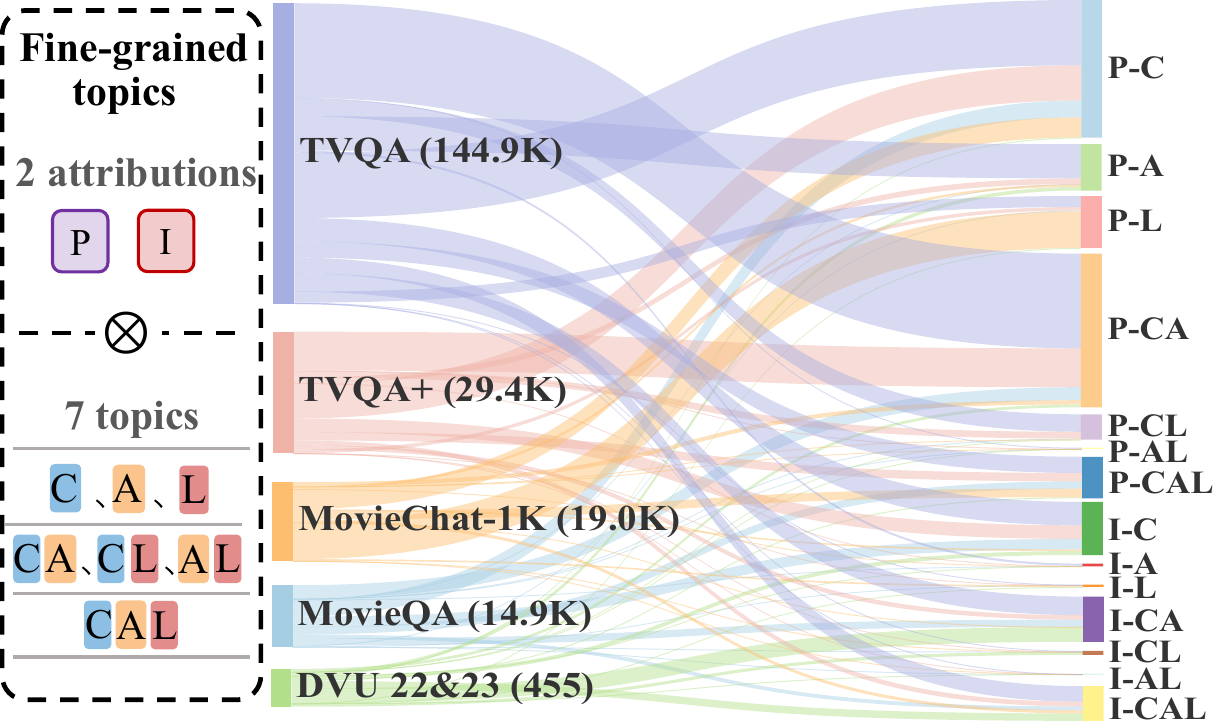}
\caption{The distribution of 5 datasets questions across the 14 fine-grained topics, based on 7 topics (character (C), action (A), location (L) and their combinations) and 2 cognitive attributions, i.e., perception (P) and inference (I).}
\label{fig:sankey}
\end{figure}

\begin{table*}[]
    \centering
    \scriptsize
    \begin{tabular}{ccccc rcccc}
    \toprule
    \multirow{2}{*}{\parbox[c][1.5em][c]{1.5cm}{\centering \textbf{Dataset}}}  & 
    \multirow{2}{*}{\parbox[c][1.5em][c]{1.5cm}{\centering \textbf{Venue}}} & 
    \multicolumn{3}{c}{\textbf{Fine-grained topic distribution}} & \multicolumn{3}{c}{\textbf{Question Scale}} &
    \multirow{2}{*}{\parbox[c][1.5em][c]{1cm}{\centering \textbf{\begin{tabular}[c]{@{}c@{}}Difficulty  \\Measure \end{tabular}}}} &
    \multirow{2}{*}{\parbox[c][1.5em][c]{1cm}{\centering \textbf{\begin{tabular}[c]{@{}c@{}}Cross \\ Episodes\end{tabular}}}} \\
    \cmidrule(lr){3-5} \cmidrule(lr){6-8}
    &  & \textbf{\# Fin. Top.} & \textbf{Gin.} & \textbf{Ent.} & \textbf{\# Que.} & \textbf{Vid. Len. (s)} & \textbf{\# Que.$\times$Vid. Len. (Ks)} &  &  \\
    \midrule\midrule
    MovieQA & CVPR'16 & 6 & 0.819 & 2.713 & ~14.9K & 202.7 & ~~3,020.2 & \textcolor{red}{\ding{55}} & \textcolor{red}{\ding{55}} \\
    TVQA & EMNLP'18 & 8 & 0.821 & 2.873  & 144.9K & 76.2 & 11,041.4 & \textcolor{red}{\ding{55}} & \textcolor{red}{\ding{55}} \\
    TVQA+ & ACL'20 & 5 & 0.789 & 2.660  & 29.4K & 61.5 & ~~1,808.1& \textcolor{red}{\ding{55}} & \textcolor{red}{\ding{55}} \\
    HLVU (DVU 22\&23) & ICMR'20& 6 & 0.773 & 2.548  & 455 & 106 / 4,907& ~~1,010.5 & \textcolor{red}{\ding{55}} & \textcolor{red}{\ding{55}} \\
    DramaQA & AAAI'21 & - & - & -  & ~17.9K & 3.6 / 91.8& ~~~~~429.9 & \textbf{\textcolor{green}{\ding{51}}} & \textcolor{red}{\ding{55}} \\
    DeepMaven & EACL'23 & - & - & -  & 1K & 3,102 & ~~3,102.0& \textcolor{red}{\ding{55}} & \textcolor{red}{\ding{55}} \\
    CinePile & CVPRW'24 & - & - & -  &\textbf{ ~200K} & 160& 32,000.0 & \textcolor{red}{\ding{55}} & \textcolor{red}{\ding{55}} \\
    MovieChat-1K & CVPR'24 & 4 & 0.701 & 2.203 & 19.0K & 564& 10,716.0 & \textcolor{red}{\ding{55}} & \textcolor{red}{\ding{55}} \\
    \midrule
    \multicolumn{2}{c}{\textbf{FriendsQA (ours)}} & \textbf{14} & \textbf{0.927} & \textbf{3.794}  & 44.6K & \textbf{1,358 / 5,390}& \textbf{98,874.8} & \textbf{\textcolor{green}{\ding{51}}} & \textbf{\textcolor{green}{\ding{51}}}\\
    \bottomrule
    \end{tabular}
\caption{Comparisons of existing DVU datasets. Fine-grained topic distribution consider the number of fine-grained topics exceeding 5\% of the dataset (\# Fin. Top.) and the balance degree of topic distribution. The Gini index (Gin.) and entropy (Ent.) are employed to measure the distribution's balance. Question scale is compared by analyzing the number of questions (\# Que.), the average video length (Vid. Len.), and their product. The proposed FriendsQA dataset offers additional features, including difficulty measure and cross episodes questioning. The figures around the ``/'' represent two distinct video input lengths. }
\label{tab:Comparisons}
\end{table*}

However, VideoQA models' performance significantly declines on deep video understanding (DVU) task. For instance, VideoChat2 achieves 61.70\% accuracy on the factoid NExT-QA dataset but drops to 44.05\% on our DVU FriendsQA dataset. As illustrated in Figure \ref{fig:example}, the key reasons include: First, regarding question attributes, in addition to perception questions that inquire about visual cues, DVU also includes inference questions that assess the understanding of storylines, offering more diverse questions. Second, for the same perception questions, factoid VideoQA involves short-range shot-level perception in factoid video depicting simple events without complex narrative interactions, whereas DVU involves long-range scene-level \cite{cliang1, bnpt23} perception in story videos encompass intricate storylines. In addition, characters and locations in DVU have specific identities \cite{cliang2}, i.e., Ross and Monica's apartment. Due to these factors, the DVU task is more challenging. The core of DVU task lies in understanding storylines with long-range evolvement, composed of story topics \cite{3w}, i.e. characters, actions, locations and their combinations.

Although story topics are essential to analyzing storylines, the most of existing DVU datasets \cite{tvqa,movieqa,moviechat} do not organize questions based on them. We use Gemini 1.5 Pro\footnote{https://aistudio.google.com/} to categorize questions from 5 classic DVU datasets (Please refer to Appendix A for more details.) into 14 fine-grained topics we refer to as in this paper, i.e., 7 story topics, character, action, location and their combinations attached to 2 question attributions, perception and inference \cite{storyPI}. 
The results are shown in Figure \ref{fig:sankey}. We find that these datasets tend to concentrate on a limited type of fine-grained topics. This situation make it difficult to provide a comprehensive evaluation of VideoQA models' DVU capability.

Furthermore, complex storylines involve long-range evolvement, necessitating both numerous questions to cover the entire storyline and sufficiently video length to convey the storyline clearly. However, as shown in Table \ref{tab:Comparisons}, existing DVU datasets either pair a large number of questions with short videos, such as TVQA \cite{tvqa}, or feature longer videos with fewer questions, such as DeepMaven \cite{deepmaven}. This limitation prevents them from evaluating the model’s understanding of storylines in both breadth and depth simultaneously.

To address the above two challenges, we propose StoryMind, a large language model (LLM) based multi-agent collaboration framework, comprising a generator to generate questions with 14 fine-grained topics and two reviewers to remove low-quality ones. We provide the generator with detailed topic explanations and manually constructed examples to guide question generation. To ensure balanced topic coverage, generator will iteratively generate questions until the number of questions for each fine-grained topic reaches the predetermined threshold. To ensure dataset quality, only the question and its answer unanimously deemed reasonable and accurate by both reviewers will be retained. Moreover, as illustrated in Table \ref{tab:Comparisons}, we perform difficulty measure for each question to accurately evaluate the DVU capability of various VideoQA models, and introduce cross-episode questions to augment the challenge of the dataset by covering more complex storyline. 

We apply StoryMind to the popular sitcom \textit{Friends}, comprising 234 episodes with average video length of 1,358s across ten seasons. This results in a new large-scale DVU dataset, FriendsQA, with the most diverse and balanced fine-grained topics. As studied in Table \ref{tab:Comparisons}, this dataset includes 44.6K questions evenly distributed across 14 fine-grained topics. Thereafter, we comprehensively evaluate the DVU capbility of 10 state-of-the-art (SOTA) models on FriendsQA, encompassing both VLM-based methods and newly proposed MLLM-based methods. 

\begin{figure*}[t]
\centering
\includegraphics[width=\linewidth]{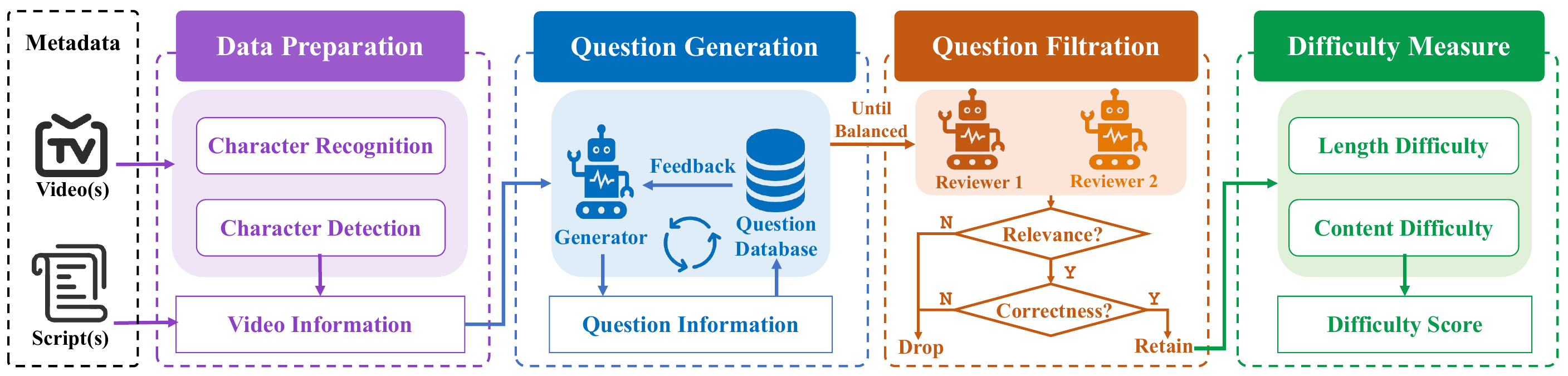}
\caption{The workflow diagram of the multi-agent collaboration framework StoryMind.}
\label{fig:arch}
\end{figure*}

In summary, our contributions are three-fold:
\begin{itemize}
    \item We devise StoryMind, a LLM-based multi-agent collaboration framework to automatically generate and review large-scale questions with diverse fine-grained topics.
    \item We construct FriendsQA, a large-scale DVU dataset comprising 44.6K questions with 14 fine-grained topics tailored for thorough DVU evaluation 
    \item We conduct extensive evaluations of 10 SOTA VideoQA models on FriendsQA.
\end{itemize}

\section{Related Work}
In this section, we briefly overview the VideoQA datasets, encompassing factoid VideoQA and DVU datasets.

\subsection{Factoid VideoQA Datasets}
Factoid VideoQA dataset \cite{msvdqa, msrvtt-mc,KnowIT-VQA, how2qa, justask21, egoschema} mainly focus on simple visual fact in short-range, such as object recognition, action recognition, spatial and temporal understanding in shot level. ActivityNet-QA \cite{activitynet-qa} focuses on actions recognition, spatial relationships, and temporal relationships. It contains 58K human-annotated QA pairs on 5.8K videos from ActivityNet \cite{activitynet}. NExT-QA \cite{nextqa} delves videos featuring object interaction, and there are 52K manually annotated questions including temporal, and descriptive questions. A notable distinction of these datasets is that they focus on short-range understanding of video with out complex storyline. This is a major factor contributing to the significant performance gap between factoid VideoQA and DVU.

\subsection{DVU Datasets}
Story videos, such as TV shows, are composed of complex interactions and long-range evolvement of core story topics. Understanding these topics requires models to possess deep video understanding (DVU) capability. Early work, such as PororoQA \cite{deepstory}, focused on scene-dialog storytelling without complex storylines. Recent research has increasingly shifted towards understanding complex storylines in story video. However, as illustrated in Figure \ref{fig:sankey}, the majority of current DVU datasets \cite{movieqa, tvqa, tvqaplus, moviechat,hlvu} do not organize questions based on story topics. This phenomenon indicates existing datasets are difficult to provide a comprehensive evaluation of the VideoQA models’ capability of DVU. 

In addition, As shown in Table \ref{tab:Comparisons}, some DVU datasets contain a large number of questions but are paired with short videos less than 5 minute, e.g., TVQA \cite{tvqa}. Conversely, the other datasets encompass long videos (approximately 90 minutes). e.g. HLVU \cite{hlvu}. Recently, the DVU 2022 \cite{dvu22} and DVU 2023 \cite{dvu23} grand challenges\footnote{https://sites.google.com/view/dvuchallenge2023/home} have comprehensively addressed on HLVU dataset. Nevertheless, These datasets often have a limited number of questions. For example, HLVU and DeepMaven \cite{deepmaven} have fewer than 1K questions, and the largest, MovieChat-1K \cite{moviechat}, has only 19K questions. It hinders the evaluation of the model's understanding of storylines in terms of both breadth and depth simultaneously.

Particularly, There are some movie/TV-based datasets \cite{friendsqa19,challenging,summscreen} for story understanding in the NLP domain. Specifically, compared to script-based QA dataset also named FriendsQA \cite{friendsqa19}, ours has two key differences: (1) In the question generation phrase, we explicitly add spatio-temporal information, such as timeline and character detection boxes obtained from video, to better suit the DVU task. (2) In the question answering phase, we provide video with QA pair as the basis of answering questions.
\begin{figure}[t]
    \centering
    \includegraphics[width=\linewidth]{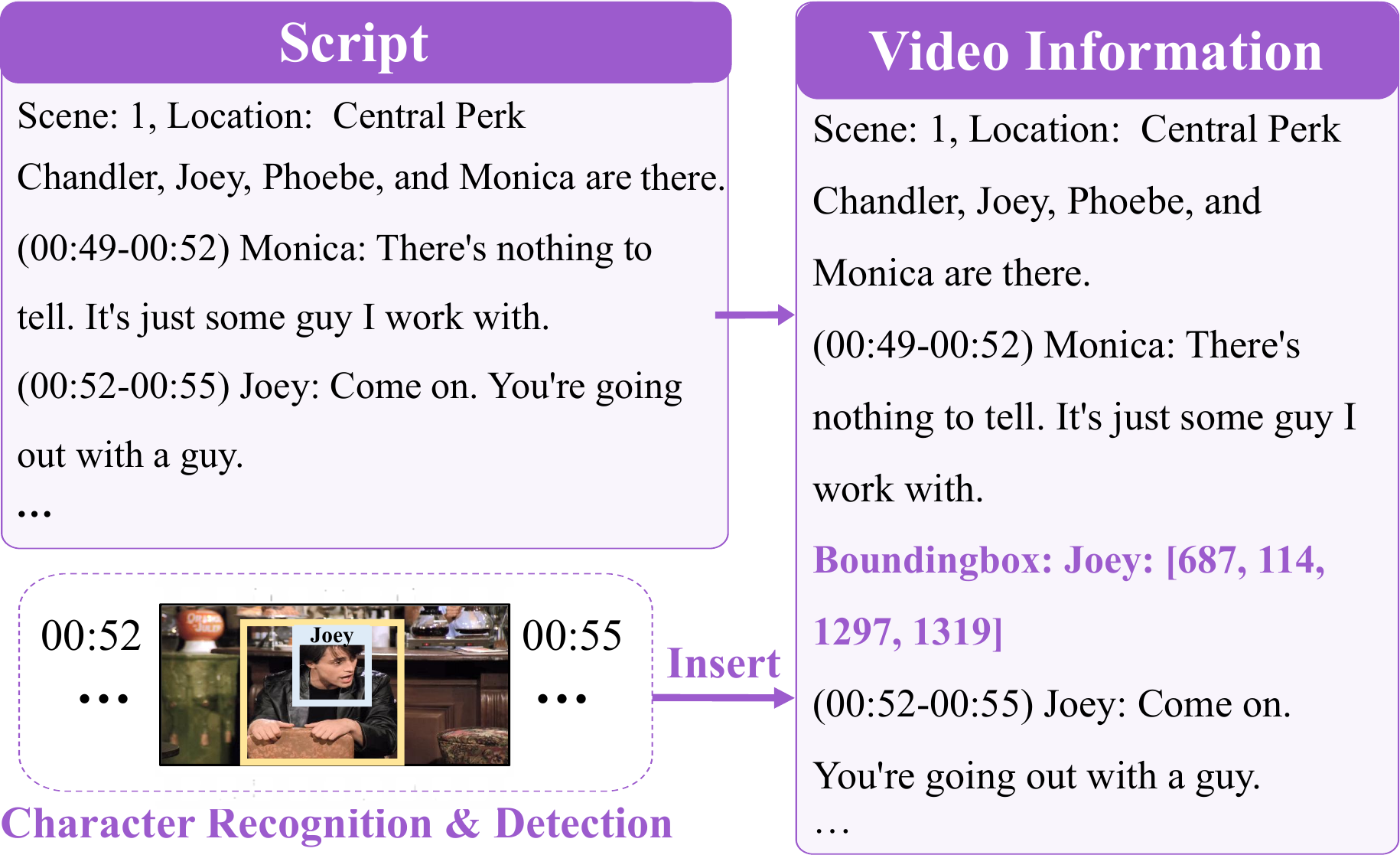}
    \caption{The flowchart of data preparation.}
    \label{fig:data_pre}
\end{figure}

\section{StoryMind}
To reduce the labor costs, we propose StoryMind, a multi-agent collaboration framework. It can automatically generate large-scale questions with conprehensive and balanced fine-grained topics. As shown in Figure \ref{fig:arch}, it mainly consists of four stages, data preparation, question generation, question filtration and difficulty measure.

\subsection{Data Preparation}
The data preparation phase is responsible for converting the metadata into video information.
\subsubsection{Metadata}
The script and episode video represent the textual structure and visual presentation of the storyline, respectively. Therefore, we use them as the metadata for question generation. The script from PAINS dataset \cite{TVCSINS} contains multiple scenes, with each scene detailing the location, the main characters involved, and aligning the dialogue with the video timeline, as shown in Figure \ref{fig:data_pre}. It provides the generator with a complete storyline evolvement ensuring that the generated questions are more accurate and closely aligned with the storyline. The timeline allows the generator to accurately generate the start and end timestamps corresponding to each question, facilitating the subsequent calculation of difficulty scores.

\subsubsection{Video Information}
 We apply the shot-based instance search method \cite{MM23} to detect and identify characters in the episode with, obtaining detection bounding boxes. This additional visual information allows the generator to generate questions with positional details, i.e., left, and right, thereby increasing questions complexity. Finally, we insert the bounding boxes into the script according to the timeline, as shown in Figure \ref{fig:data_pre}, creating a crucial component of the question generation prompts, which we refer to as the \textit{video information}. Notebly, to enquire the more complicated storylines, we specifically designed cross-episode questions by concatenating 4 consecutive episodes\footnote{We choose 4 episodes because the scripts from 4 episodes reach the generator's context limit.}. 

\begin{figure}[t]
\centering
\includegraphics[width=\linewidth]{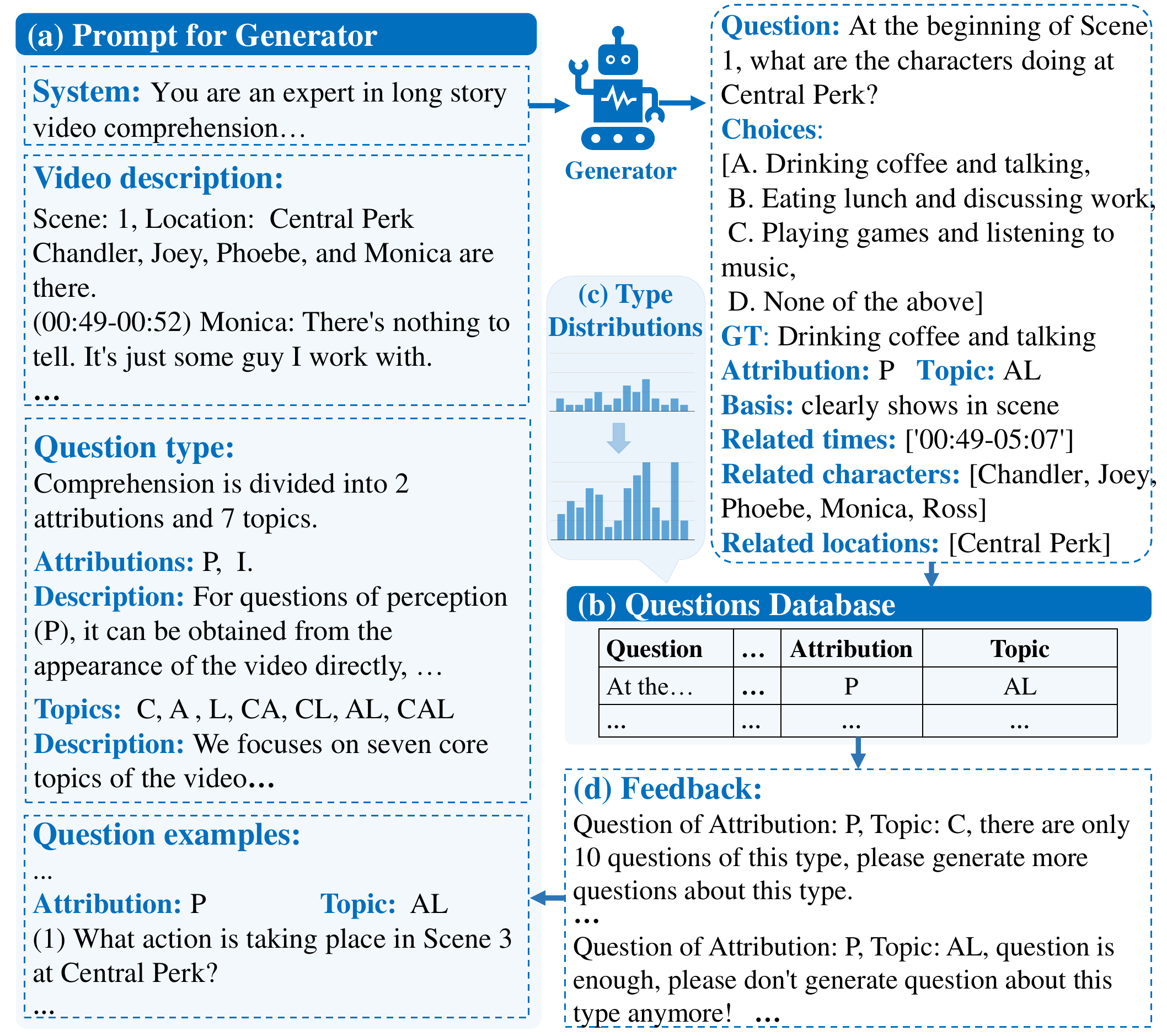}
\caption{The iterative flowchart of question generation.}
\label{fig:qg}
\end{figure}

\subsection{Question Generation}
The question generation stage aims to produce questions with diverse and balanced fine-grained topics. 
We introduce Gemini 1.5 Pro as generator and prompt the generator with \textit{video information}, \textit{descriptions of the fine-grained topics} and \textit{questions examples}, as illustrated in Figure \ref{fig:qg}(a). Video information is the output of data preparation stage that record the script, timeline, and character positions. Descriptions of the fine-grained topic contains the description of 2 attributions (P, I) and 7 topics (C, A, L, CA, CL, AL and CAL). Questions examples include manually designed examples for each fine-grained topic to assist the generator in better understanding the fine-grained topics. ALL of them help the generator understand the categorization of fine-grained topics and generate corresponding questions.

In addition, to ensure a balanced distribution of questions across each fine-grained topic, we require the generator to specify the fine-grained topic categorization of each question and save the information into question database (Figure \ref{fig:qg}(b)). This allows database to track the distribution across all fine-grained topics (Figure \ref{fig:qg}(c)). To prevent the generator from focusing solely on one topic, we devise a feedback mechanism. The database checks the number of questions for each fine-grained topic against a same threshold. If a topic falls below the threshold, the database will provide feedback to generator with generating more questions for that topic. Otherwise, the opposite feedback will be given (Figure \ref{fig:qg}(d)). The generator will iteratively generate questions until all topics reach the threshold.

As illustrated in generated result of Figure \ref{fig:qg}, we require the generator to output additional information for each question, such as choices list and ground truth (GT). Since the video information contains a complete timeline, character and location information, we require the generator to specify the related time, characters, and locations associated with each question. These will be used for subsequent difficulty measure. Please refer to Appendix B.1 for more details. 

\begin{figure}[t]
    \centering
    \includegraphics[width=\linewidth]{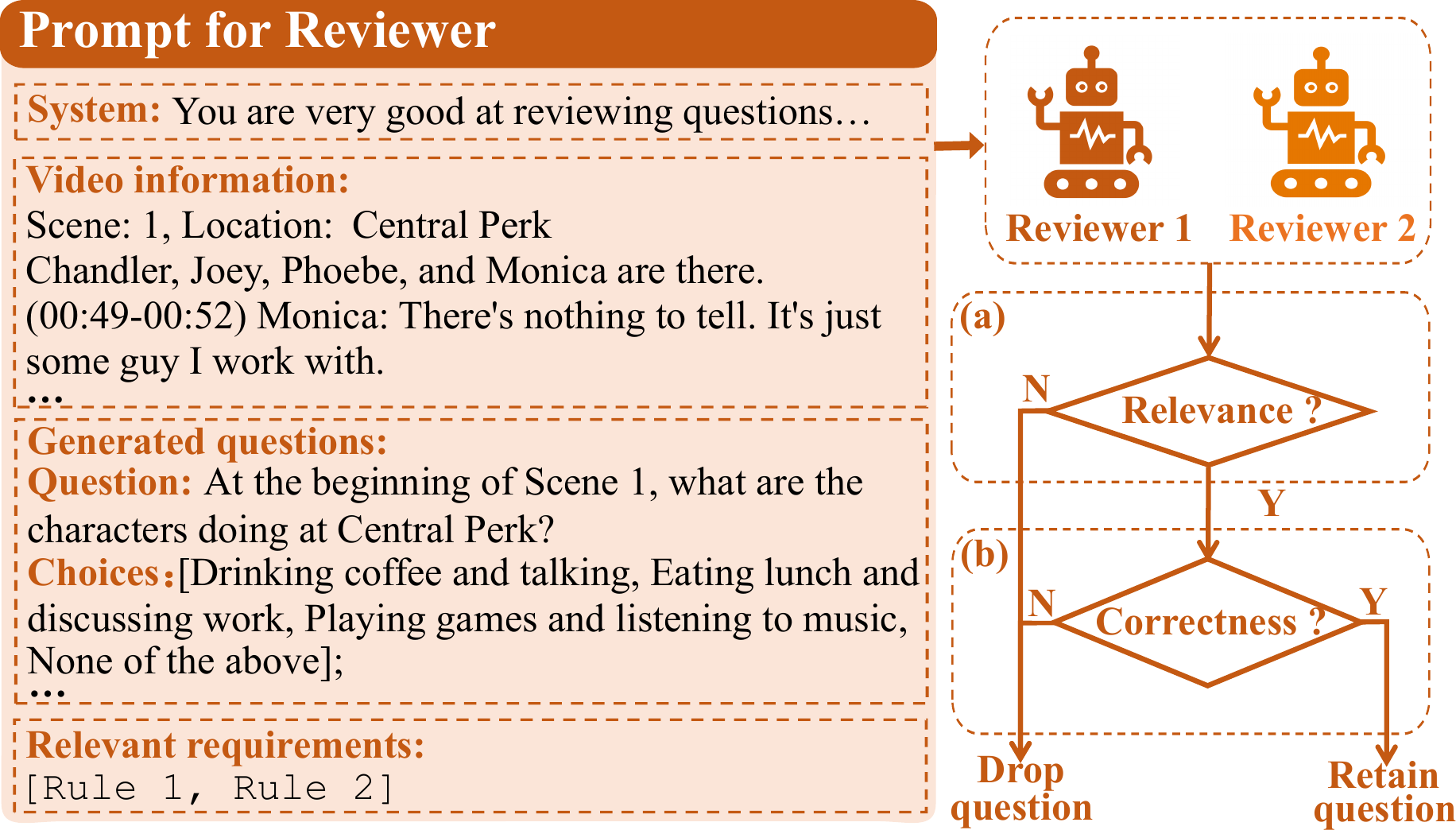}
    \caption{The flowchart of question filtration.}
    \label{fig:reviewer}
\end{figure}

\subsection{Question Filtration}
Question filtration aims to filter out incorrect questions. We introduce Gemini 1.5 Pro and Claude 3.5 Sonnet\footnote{https://claude.ai/} as two reviewers to automate the process. The prompt is meticulously designed for them as shown in Figure \ref{fig:reviewer} (For complete prompt, please refer to Appendix B.2), which mainly consists of three parts, \textit{video information}, \textit{generated questions} and \textit{relevant requirements}. Video information is the output of data preparation. It ensures reviews to be on the same level of obtaining storylines as the generator, allowing them to fairly assess the questions' accuracy. Generated questions is the output of question generation phase. Reviewers evaluate the generated questions based on the video information. Relevant requirements include two rules:

\begin{itemize}
    \item \textbf{Rule 1} The question must be relevant to the video and can be answered with GT from the generator. 
    \item \textbf{Rule 2} Among the 4 choices, there must be only one correct answer, and three wrong answers. 
\end{itemize}
\begin{figure}[t]
\centering
\includegraphics[width=\linewidth]{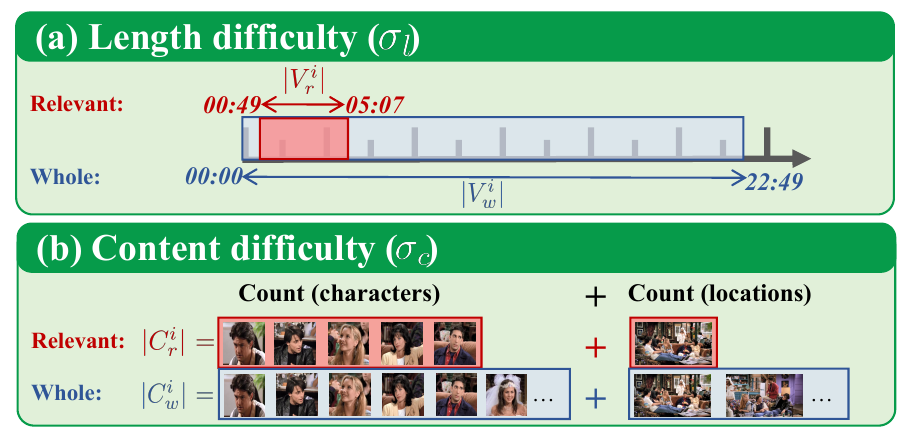}
\caption{Illustration of two difficulty factors.}
\label{fig:diff}
\end{figure}

If the both rules are met, the output for the correctness of the question is \texttt{True}; otherwise, it is \texttt{False} (Figure \ref{fig:reviewer}(a)).  Additionally, The unique correct answer selected independently by each of the two reviewers must be identical and consistent with GT (Figure \ref{fig:reviewer}(b)).

\subsection{Difficulty Measure}
Difficulty measure aims to assign a difficulty score to each question to better assess the DVU capabilities of VideoQA models \cite{nju22}. We propose 2 difficulty factors, i.e., video length and content. Let $\vert V^i_w \vert$ and $\vert V^i_r \vert$ denote the length of the whole video and relevant video for $i$-th question. $\vert C^i_w \vert$ and $\vert C^i_r \vert$ denote the instance (including characters and locations obtained from video information) number of the whole video and relevant video to represent content. Therefore, we define 2 difficulty scores:
\begin{itemize}
    \item Length difficulty ($\sigma_l^i= \vert V^i_w \vert/\vert V^i_r \vert$). It represent the relationship of length between whole video and relevant video as shown in Figure \ref{fig:diff}(a). Larger value indicates smaller proportion of relevant video length, making it harder for models to capture the necessary information.
    \item Content difficulty ($\sigma_c^i= \vert C^i_w \vert/\vert C^i_r \vert$). It characterizes the relationship of video content between whole video and relevant video as shown in Figure \ref{fig:diff}(b). Larger value indicates less video content of relevant video, making it harder for models to capture the necessary instance.
\end{itemize}

Based on the two factors mentioned above, we use the following formula to calculate the overall difficulty score:
\begin{equation}
    \sigma^i = \frac{\sigma^i_l}{\mu_l}+\frac{\sigma^i_c}{\mu_c}
\end{equation}
where $\mu_l$ and $\mu_c$ represent the average length and content scores for all questions, which is used to mitigate factor discrepancy. In the ``Evaluation'' section, we validate the effectiveness of our difficulty measure through experiments.

\begin{table}[]
    \centering
    \fontsize{7pt}{\baselineskip}\selectfont
    \setlength{\tabcolsep}{1mm}{
        \begin{tabular}{ccccccc}
        \toprule
        \multirow{2}{*}{{\centering\textbf{Dataset}}} & \multirow{2}{*}{{\centering\textbf{Manul}}} & \multirow{2}{*}{{\centering\textbf{Revision}}} & \multicolumn{2}{c}{\textbf{Single}} & \multicolumn{2}{c}{\textbf{Cross}} \\

        \cmidrule(lr){4-5} \cmidrule(lr){6-7}
        & & &\textbf{\# Num} &\textbf{Ratio} &\textbf{\# Num} &\textbf{Ratio}\\
        \midrule \midrule
        FriendsQA-S1 & \textcolor{red}{\ding{55}} & \textcolor{red}{\ding{55}} & 3,795 & 100\%  & 995 & 100\% \\
        FriendsQA-M w/o R. & \textcolor{green}{\ding{51}} & \textcolor{red}{\ding{55}}  & 3,475 & 91.57\% & 894 & 89.85\% \\
        FriendsQA-M & \textcolor{green}{\ding{51}} & \textcolor{green}{\ding{51}}  & 3,584 & 94.44\% & 905 & 90.95\% \\

        \bottomrule
        \end{tabular}%
    }
    \caption{Comparison of datasets on first season in different filtration and type. w/o R. indicates revison are not included.}
    \label{effectiveness of automatic}
\end{table}

\begin{table}[t]
    \centering
    \fontsize{9pt}{\baselineskip}\selectfont
    \centerline{
        \begin{tabular}{cccc}
        \toprule
        \multirow{2}{*}{{\centering\textbf{Type}}} & \multicolumn{2}{c}{\textbf{Dataset}} &\multirow{2}{*}{{\centering\textbf{Difference}}}\\
        \cmidrule(lr){2-3}
        & \textbf{FriendsQA-S1} & \textbf{FriendsQA-M}\\
        \midrule \midrule
        single & 33.45 & 33.60 & 0.15\\
        cross & 34.77 & 35.02 & 0.25 \\
        \bottomrule
        \end{tabular}%
    }
    \caption{The average accuracy and difference (\%) of 10 SOTA models across single and cross-episode questions on FriendsQA-M and FriendsQA-S1.}
    \label{tab:friendsqa-M-S1}
\end{table}


\section{FriendsQA Dataset}
\subsection{Dataset Quality}
\subsubsection{Manual Verification}
Before presenting the dataset statistics, we assess the quality of the FriendsQA generated by the StoryMind. This evaluation ensures the validity of subsequent evaluation.
Therefore, we first select 4,790 questions from the first season (denoted as FriendsQA-S1) and apply the same requirements as in the previous ``Question Filtration'' and manually revise questions with incorrect options (e.g., the correct answer is A but B is selected). In the subsequent experiments, we use the manually verified dataset with revision (denoted as FriendsQA-M). 

\subsubsection{Manual v.s. Automatic}
The comparisons between FriendsQA-M and FriendsQA-S1 are shown in Table \ref{effectiveness of automatic}. For single-episode questions, 91.57\% are retained directly without revision (denoted as FriendsQA w/o R.), and 94.44\% are retained with revision. For cross-episode questions, these figures are 89.85\% and 90.95\%, respectively. Furthermore, we conduct analysis of the accuracy difference of 10 SOTA models (will be elaborated in the latter ``Evaluation'' section) on FriendsQA-S1 and FirendsQA-M, as shown in Table \ref{tab:friendsqa-M-S1}. The average differences between them are 0.15\% for single-episode questions and 0.25\% for cross-episode questions. These results demonstrate the high-quality of the FriendsQA and the effectiveness of the StoryMind.

\subsection{Dataset Statistics}
\subsubsection{Dataset Scale}
We applied StoryMind to the classic sitcom \textit{Friends}, which contains 234 episodes with average length of 1358s. It leads to a large-scale dataset, FriendsQA, with over 44.6K questions, covering 14 fine-grained topics. The number of single-episode and cross-episode questions are 35,222 and 9,470, respectively, in approximately a 4:1 ratio. More examples for 14 fine-grained topic are elaborated in Appendix C. 
\subsubsection{Fine-grained Topic}
\begin{figure}
    \centering
    \includegraphics[width=\linewidth]{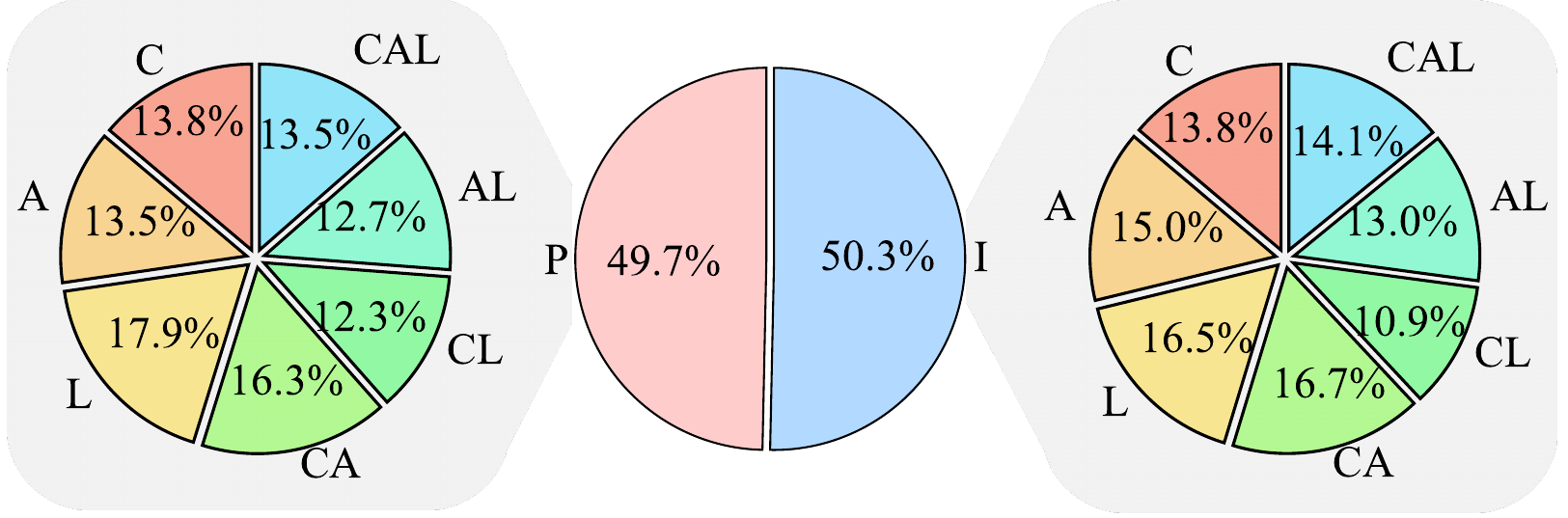}
    \caption{The distribution of fine-grained topics.}
    \label{fig:distribution}
\end{figure}

As illustrated in Figure \ref{fig:distribution}, The proportion of P and I questions is nearly 50\%, and for specific topics, like the CL with the lowest proportion of I questions, it still accounts for 10.9\%. This indicates that the questions are evenly distributed across the 14 fine-grained topics. The distribution is the same for single and cross-episode question, which is detailed in Appendix D.

\subsubsection{Difficulty}
\begin{figure}[t]
    \centering
    \includegraphics[width=\linewidth]{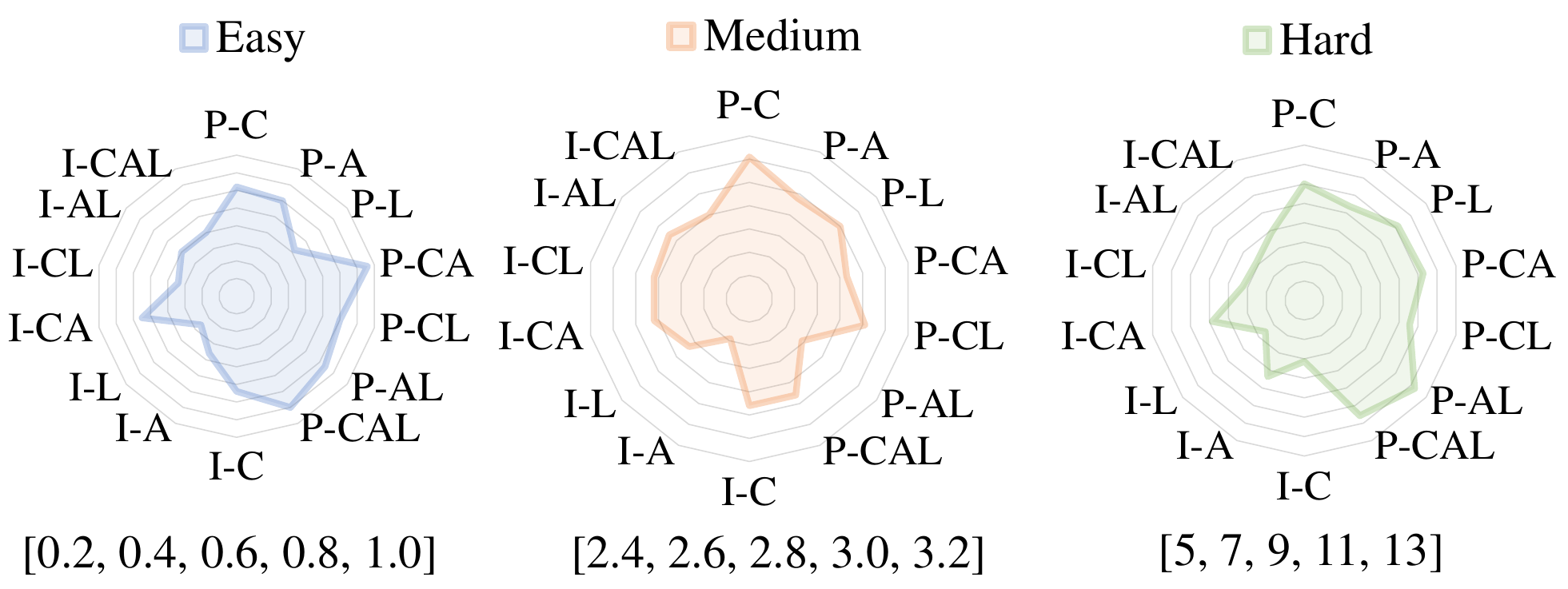}
    \caption{Overall difficulty score on different levels. The numbers in [$\cdot$] are coordinate values for each circle.}
    \label{fig:difficulty_level}
\end{figure}

Finally, to facilitate the analysis of results in conjunction with question difficulty, 
we sort the questions in ascending order of difficulty score $\sigma^i$ and divide them into three levels, easy, medium and hard based on a 9:3:1 ratio. The average difficulty scores of the 14 fine-grained topics across different difficulty levels are shown in the Figure \ref{fig:difficulty_level}. It can be observed that the difficulty of P questions tends to be higher than that of I questions across all three levels. 

\section{Evaluation}
\subsection{Setting}
\subsubsection{Baseline}
We benchmark 10 SOTA models over the past two years and categorize these models into two groups:
\begin{itemize}
    \item \textbf{MLLM} models including  Chat-Univi \cite{chat-univi}, MA-LMM \cite{malmm}, MovieChat \cite{moviechat}, SeViLA \cite{sevila}, TimeChat \cite{timechat}, Video-ChatGPT \cite{videochatgpt}, VideoChat2 \cite{mvbench}, VideoLLaMA2\cite{videollama2}. Except for SeViLA, which uses FlanT5-XL 3B \cite{flant5} as backbone, all other MLLM models, use various versions of the LLaMA model \cite{llama,llama2} as backbone. For these models, we consistently use the 7B parameter version of LLaMA as the backbone. 
    \item \textbf{VLM} models including  Vid-TLDR \cite{vid-tldr} and VIOLETv2 \cite{violetv2}. We use the weights fine-tuned on MSRVTT-QA, the most commonly used VideoQA dataset, to ensure fairness.
\end{itemize}

\begin{table*}[t]
    \centering
    \scriptsize
    \setlength{\tabcolsep}{0.7mm}{
    \begin{tabular}{ccccccccccccc}
        \toprule
        \multirow{2}{*}{\centering\textbf{Att.}} & \multirow{2}{*}{\centering\textbf{Top.}} & \multicolumn{8}{c}{\textbf{MLLM}}                                                                                                                                                                                                                                                                                                                                                                                                                                               & \multicolumn{2}{c}{\textbf{VLM}}      & \multirow{2}{*}{\centering\textbf{AVG}} \\  \cmidrule(lr){3-10} \cmidrule(lr){11-12}
                                            &                                 & \textbf{Chat-UniVi}                        & \textbf{MA-LMM}                                                               & \textbf{MovieChat} & \textbf{SeViLA}                                                               & \textbf{TimeChat} & \textbf{Video-ChatGPT} & \textbf{VideoChat2}                                                           & \textbf{VideoLLaMA2}                                                          & \textbf{VIOLETv2} & \textbf{Vid-TLDR} &                               \\  \midrule \midrule
        \multirow{7}{*}{P}                  & C                               & 26.27 / 23.85                              & 29.84 / 35.90 & 17.76 / 21.24      & \textbf{35.87} / 40.75 & 18.07 / 15.28     & 15.82 / 17.52          & 23.53 / 30.31                                                                 & 28.87 / \textbf{44.47} & 26.71 / 22.11     & 21.33 / 22.61     & 24.41 / 27.40                 \\
                                            & A                               & 25.77 / 41.79                              & 30.47 / 36.48 & 24.51 / 22.06      & \textbf{39.19}/ 27.03                                  & 19.26 / 17.41     & 25.39 / 20.07          & 22.70 / 34.16                                                                 & 32.10 / \textbf{42.45} & 27.07 / 28.86     & 22.53 / 26.87     & 26.90 / 29.72                 \\
                                            & L                               & 24.77 / 34.62                              & 32.79 / 43.24 & 17.56 / 19.32      & 27.79 / 34.03                                                                 & 21.53 / ~~5.65    & 15.41 / 16.05          & 29.19 / 34.77 & \textbf{37.54} / \textbf{45.32} & 28.52 / 25.56     & 24.07 / 27.04     & 25.92 / 28.56                 \\
                                            & CA                              & 30.77 / 30.84                              & 29.42 / 31.95 & 16.07 / 22.03      & 28.38 / 33.29                                 & 22.71 / 14.32     & 15.07 / 15.79          & 23.78 / 27.29                                                                 & \textbf{33.27} / \textbf{35.37} & 25.53 / 28.40     & 24.38 / 23.75     & 24.94 / 26.30                 \\
                                            & CL                              & 27.43 / 27.41                              & 31.05 / \textbf{42.27} & 19.00 / 18.99      & \textbf{34.96} / 31.24                                 & 14.46 / ~~5.51    & 16.73 / 12.10          & 27.34 / 36.75                                 & 32.55 / 37.67 & 28.16 / 27.72     & 21.26 / 24.50     & 25.29 / 26.42                 \\
                                            & AL                              & 31.58 / 40.70                              & 29.77 / \textbf{47.51}                                 & 21.95 / 22.76      & 34.39 / 31.56                                 & 18.46 / ~~6.48    & 24.25 / 18.94          & 24.71 / 38.70                                                                 & \textbf{34.80} / 45.68 & 28.96 / 29.07     & 25.34 / 27.24     & 27.42 / 30.86                 \\
                                            & CAL                             & 30.74 / 27.62                              & 29.94 / \textbf{41.77}                                 & 19.86 / 22.31      & 31.88 / 37.14 & 19.68 / ~~7.35    & 16.12 / 13.20          & 25.36 / 29.93                                                                 & \textbf{35.01} / 37.14 & 28.36 / 30.20     & 25.67 / 23.27     & 26.26 / 26.99                 \\ \midrule
        \multirow{7}{*}{I}                  & C                               & 38.62 / 30.03                                                                 & 49.16 / 42.90                                 & 23.24 / 23.43      & 36.10 / 47.36                                & 31.79 / 13.53     & 25.40 / 22.77          & 39.02 / 43.56 & \textbf{52.24} / \textbf{57.43} & 28.83 / 31.52     & 25.36 / 26.73     & 34.98 / 33.93                 \\
                                            & A                               & 57.69 / 72.14                              & 54.67 / 50.88                                                                 & 25.60 / 27.57      & 34.50 / 25.81                                                                 & 33.46 / 27.86     & 33.72 / 42.08          & 55.86 / 76.10 & \textbf{65.57} / \textbf{82.26} & 26.94 / 24.05     & 29.81 / 35.63     & 41.78 / 46.44                 \\
                                            & L                               & 56.91 / 69.04                              & 59.62 / 55.88                                 & 26.68 / 32.04      & 32.94 / 32.20                                                                 & 41.81 / 24.15     & 32.91 / 44.12          & \textbf{64.32 / 74.46}                                   & \textbf{73.03} / \textbf{86.22} & 29.81 / 30.50     & 33.27 / 42.57     & 45.13 / 49.12                 \\
                                            & CA                              & 42.23 / 46.37                              & 43.93 / 47.72 & 20.89 / 26.45      & 32.48 / 34.32                                                                 & 33.23 / 15.87     & 26.04 / 27.55          & 38.00 / 44.53                                                                 & \textbf{52.11} / \textbf{61.38} & 30.44 / 33.21     & 27.06 / 30.75     & 34.64 / 36.81                 \\
                                            & CL                              & 39.28 / 56.08                              & 47.29/ 55.57                                  & 26.33 / 34.80      & 32.40 / 34.63                                                                 & 15.64 / 18.75     & 27.19 / 43.24          & 44.22 / 67.91 & \textbf{49.38} / \textbf{75.84} & 39.23 / 37.16     & 26.76 / 40.88     & 34.77 / 46.49                 \\
                                            & AL                              & 50.32 / 65.86                              & 49.98 / 45.06                                  & 25.63 / 30.33      & 35.21 / 29.29                                                                 & 26.90 / 22.53     & 29.27 / 36.92          & 48.92 / 66.72                                 & \textbf{59.48} / \textbf{74.18} & 32.84 / 25.13     & 27.32 / 44.02     & 38.59 / 44.00                 \\
                                            & CAL                             & 51.24 / 54.95                              & 51.24 / 39.34                                 & 24.70 / 28.83      & 33.17 / 29.58                                                                 & 29.90 / 18.32     & 34.37 / 33.18          & 53.44 / 53.90 & \textbf{64.27} / \textbf{63.81} & 33.69 / 26.73     & 31.85 / 35.74     & 40.79 / 38.44                 \\ \midrule
        \multicolumn{2}{c}{AVG}    &38.48  / 43.62   &  41.02 / 43.69  &  22.03 / 24.96      &  33.29 / 33.58   &  25.54 / 15.12    & 24.18 / 25.48           &  37.68 / 46.17                                 & \textbf{47.12} / \textbf{55.64}  &  29.46 / 28.50      &  26.32 / 30.40     & 32.51 / 34.72      \\
        \bottomrule
    \end{tabular}
    }
\caption{The accuracy (\%) of 10 SOTA models on FriendsQA. The figures before the ``/'' denote single-episode questions, while those after it denote cross-episode questions. Here the bold indicates the best value. Att. means attribution, Top. means topic.}
\label{tab:topic_acc}
\end{table*}

\begin{figure}[t]
\centering
\includegraphics[width=\linewidth]{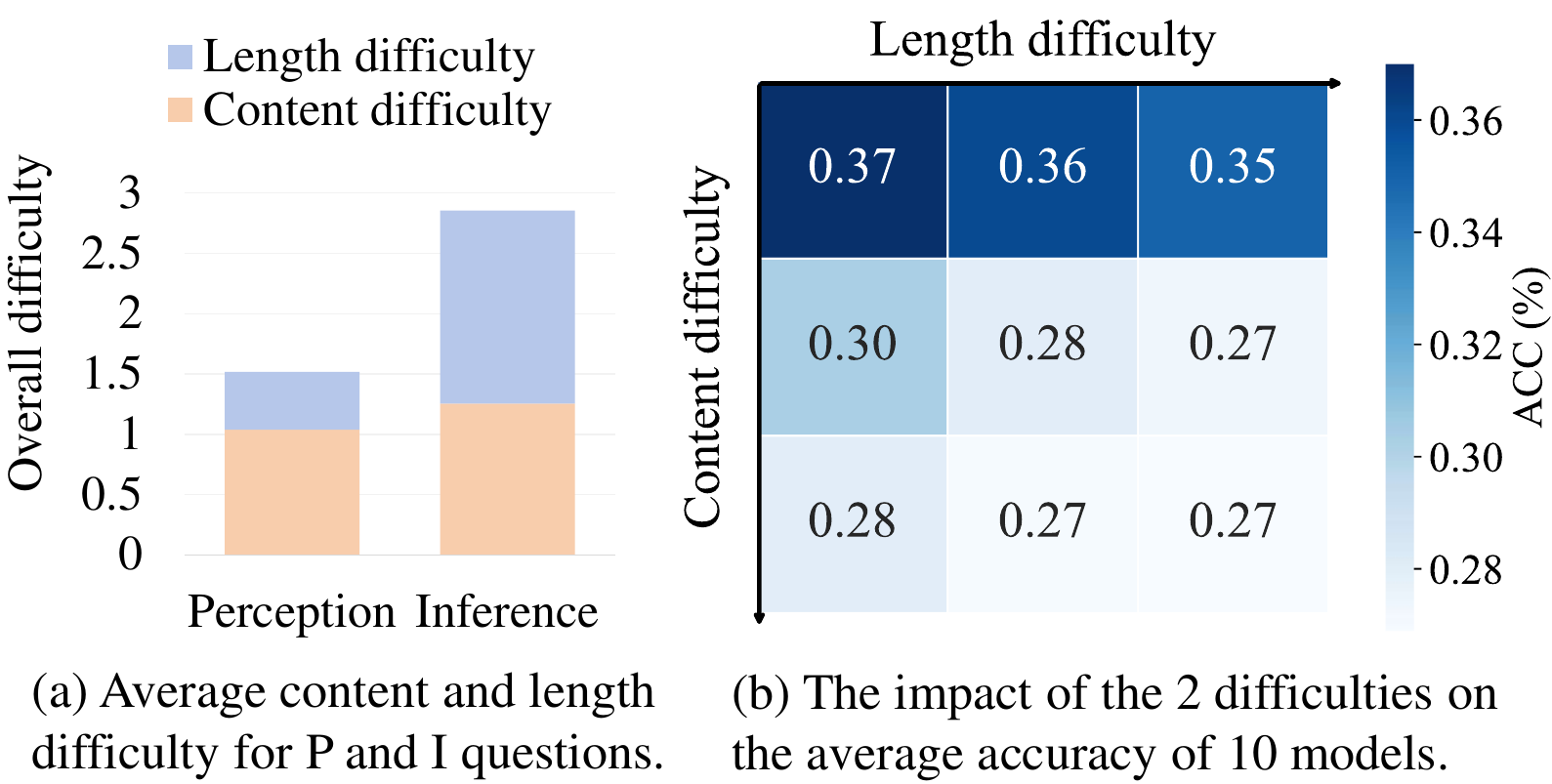} 
\caption{The influence of the two difficulties on accuracy}
\label{fig:2Difficulty}
\end{figure}

\subsubsection{Implementation Details}
We set models in zero-shot question-answering settings on FriendsQA, with official default configurations.  Evaluation is performed with the maximum frames supported by each model while staying within 48GB capacity of RTX A6000 GPU. More implementation details of 10 tested models are elaborated in Appendix E.
\subsubsection{Evaluation Metrics}
We use accuracy \cite{MM23} as metrics, which is calculated by dividing the number of correct answered questions by the total number of questions.

\subsection{Evaluation Result}
\subsubsection{MLLM v.s. VLM}
Table \ref{tab:topic_acc} shows an overall result of 10 models. 
By comparing MLLM models and VLM models, we find most MLLM models achieve better performance. This may be due to the fact that MLLMs utilize LLM as backbone, having greater language comprehension ability. Furthermore, extensive language pre-training may have brought more prior knowledge of reasoning to MLLM.

\subsubsection{Attribution} 
As shown in Table \ref{tab:topic_acc} that the accuracy of P questions is significantly lower than I questions for each model. We calculate the length difficulty $\sigma_l$ and content difficulty $\sigma_c$ of both question types. Figure \ref{fig:2Difficulty}(a) shows that P questions get higher overall difficulty, consistent with the results. Specific to 2 factors, although $\sigma_c$ is similar, P questions have significantly higher $\sigma_l$. This suggests the shorter video related to P questions make it harder for the model to extract relevant information.
\begin{figure}[t]
\centering
\includegraphics[width=\linewidth]{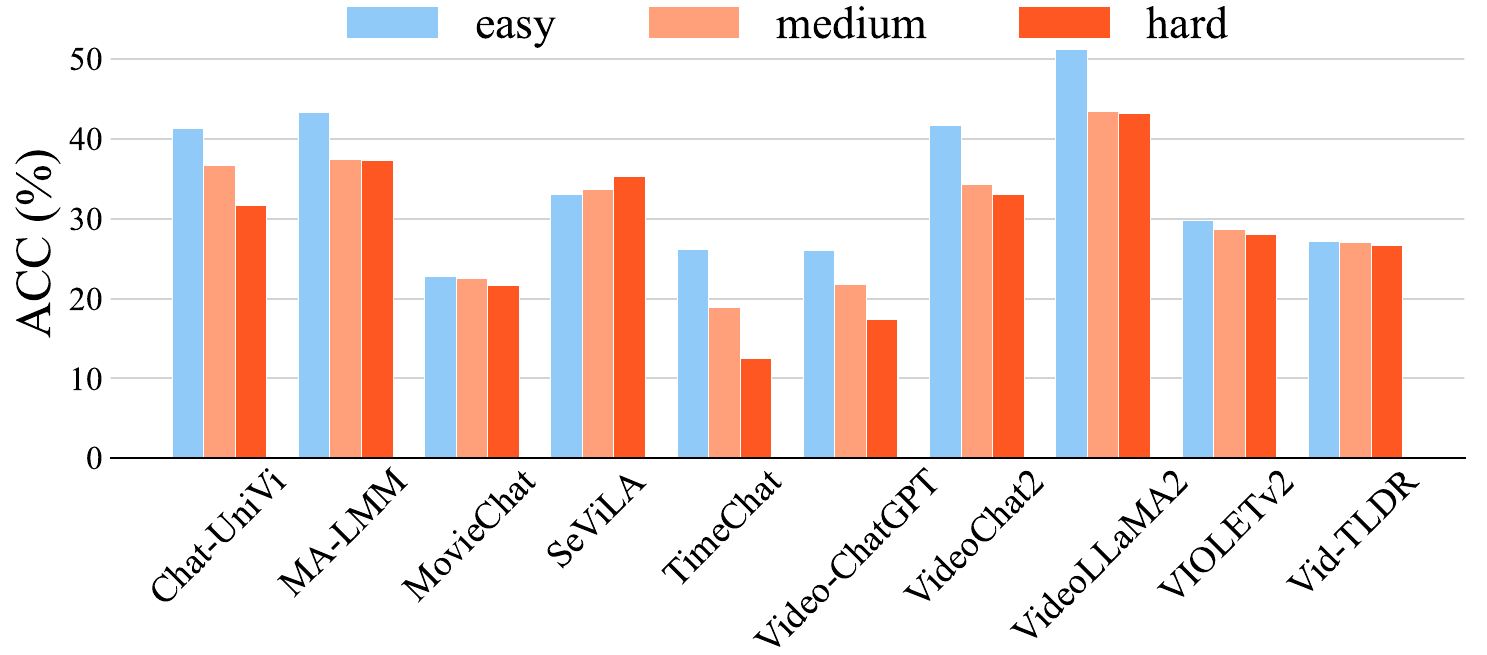} 
\caption{Accuracy (\%) of 10 SOTA models across different levels of difficulty in FriendsQA.}
\label{fig:difficulty_acc}
\end{figure}
\subsubsection{Topic} 
By Comparing the accuracy of C, A, L within P and I in Table \ref{tab:topic_acc}, We find that models exhibit lower accuracy on questions involving C. Furthermore, this trend is reinforced by comparing the accuracy of CA, CL, and AL topics within the P. This issue may stems from the model's weakness in character recognition. Unlike traditional facial recognition, in this experiment, we do not provide an additional facial registration database. The model must infer characters by integrating visual content with subtitles, making character recognition a challenging task in this context. However, this pattern is not entirely consistent for cross-episode I questions. We attribute this exception to the fact that I questions can also be answered by reasoning about actions or locations, without solely relying on character recognition. 

\subsubsection{Difficulty}
For questions difficulty, we first examine the impact of the two difficulty factors on model performance. Figure \ref{fig:2Difficulty}(b) shows the average accuracy of 10 models across 3 difficulty levels as the two difficulty scores increase. The downward trend in accuracy as difficulty rises confirms the effectiveness of these factors. Figure \ref{fig:difficulty_acc} shows the accuracy of each models in 3 difficulty levels. As observed, 9 of 10 models are consistent with our expectation, where accuracy decreases as difficulty increases. Only SeViLA shows no significant difference among different difficulty levels. It might be because SeViLA is the only model fine-tuning on TVQA including \textit{Friends}, with more prior knowledge.

\section{Conclusion and Future Work}
In this paper, we devise StoryMind, a LLM-based multi-agent collaboration framework to automatically generate large-scale dataset, FriendsQA, with fine-grained topics. FriendsQA builds upon the core C, A, and L story topics of stories with a large number of questions. 
We conduct comprehensive evaluation on FriendsQA and believe that it will guide the development of VideoQA methods.

Although FriendsQA represents a significant advancement, it remains in early stages of development. We are continually refining and expanding StoryMind framework to enhance its adaptability and applicability to various types of story videos. Inspired by \cite{wangxiao20,emnlpqg}, future work includes extending StoryMind to other genres and shows with different storytelling styles and wider range of scenarios, such as movies and dramas.

\section*{Acknowledgements}
This work is supported by the National Natural Science Foundation of China (No. 62372339, 62371350, 62372336) and the Ministry of Education Industry-University Cooperative Education Project (No. 240700006245501, 220800006041043, 202102246004, 202002142012). The numerical calculations in this paper have been done on the supercomputing system in the Supercomputing Center of Wuhan University.

\bibliography{aaai25}

\newpage
\appendix
\section{Appendix}
This appendix provides additional information not described in the paper due to the page limit. Specifically, section A contains the specific method and more quantitative results of fine-grained topic categorization. Section B discusses more details about the implementation of StoryMind, including the generator and two reviewers. Section C illustrates samples of FriendsQA. Section D represents the quantity distribution of fine-grained topic questions in FriendsQA. For section E, we elaborate on the experiment settings of the evaluation of 10 SOTA models.

\section{A. Fine-Grained Topic Categorization}
We employ Gemini 1.5 Pro to categorize fine-grained topics of questions of 5 classic DVU datasets. The prompt template for assessment is shown in Figure \ref{fig:top_cat}.

\begin{table*}[t]
\centering
\caption{Categorization results of 14 fine-grained topic categorizations in different DVU datasets.}
\label{tab:com}
\resizebox{\textwidth}{!}{%
\begin{tabular}{crrrrrrrrrrrr}
\toprule
\multirow{2}{*}{Topic} & \multicolumn{2}{c}{MovieQA} & \multicolumn{2}{c}{TVQA} & \multicolumn{2}{c}{TVQA+} & \multicolumn{2}{c}{DVU22\&23} & \multicolumn{2}{c}{MovieChat-1K} & \multicolumn{2}{c}{\textbf{FriendsQA}} \\ \cmidrule(lr){2-3} \cmidrule(lr){4-5} \cmidrule(lr){6-7} \cmidrule(lr){8-9} \cmidrule(lr){10-11} \cmidrule(lr){12-13}
 & \multicolumn{1}{c}{P} & \multicolumn{1}{c}{I} & \multicolumn{1}{c}{P} & \multicolumn{1}{c}{I} & \multicolumn{1}{c}{P} & \multicolumn{1}{c}{I} & \multicolumn{1}{c}{P} & \multicolumn{1}{c}{I} & \multicolumn{1}{c}{P} & \multicolumn{1}{c}{I} & \multicolumn{1}{c}{P} & \multicolumn{1}{c}{I} \\ 
 \midrule \midrule
C & 4,097 & 2,528 & 31,412 & 11,313 & 8,464 & 3,292 & 2 & 48 & 4,840 & 384 & 3,074 & 3,110 \\
A & 85 & 48 & 16,513 & 1,049 & 1,349 & 62 & 44 & 3 & 627 & 2 & 2,986 & 3,369 \\
L & 208 & 92 & 5,311 & 199 & 848 & 16 & 0 & 0 & 8,836 & 413 & 3,976 & 3,712 \\
CA & 3,157 & 1,746 & 45,567 & 8,934 & 9,356 & 1,099 & 34 & 179 & 1,031 & 5 & 3,618 & 3,747 \\
CL & 204 & 42 & 8,432 & 349 & 1,635 & 19 & 2 & 38 & 42 & 0 & 2,727 & 2,453 \\
AL & 7 & 2 & 462 & 42 & 70 & 5 & 0 & 3 & 7 & 0 & 2,812 & 2,934 \\
CAL & 1,792 & 936 & 7,813 & 7,517 & 2,011 & 1,157 & 13 & 89 & 2,060 & 770 & 3,006 & 3,168 \\
\bottomrule
\end{tabular}%
}
\end{table*}

As shown in Table \ref{tab:com}, results of fine-grained topic categorization indicate that the current DVU datasets are not evenly distributed. In contrast, FriendsQA has a more comprehensive and balanced distribution of various topics.

\section{B. Details for StoryMind}
In this section, we discuss more details about the implementation of StoryMind, including the prompt for the generator and reviewers.
\subsection{B.1 Prompt for Generator}
We use Gemini 1.5 Pro\footnote{https://aistudio.google.com/}, paired with LangChain\footnote{https://www.langchain.com/}, as the LLM to generate questions for the given story videos. We develop a formatting tool using Langchain's tools module, which enables the generator to output questions and related information according to our specifications. The prompt template for question generation is shown in Figure \ref{fig:generator}. 

The prompt template contains three main parts: video information, description of the fine-grained topic and question example. The first two parts have been discussed in detail in the paper. As for the final question example, we design heuristic question examples for each fine-grained topic, which enables the generator to mimic and generate new questions. Some question examples are presented in Table \ref{tab:qgexample}.

\subsection{B.2 Prompt for Reveiwers}
The structure of the prompt template for the reviewer is shown in Figure \ref{fig:reviewer_prompt}. In this prompt, each row of the CSV file corresponds to a generated question and its choices.

The first step in the reviewer's verification process is the relevance review. This step requires utilizing the video information, question, and choices provided in the prompt. The video information offers a textual description of the video content, allowing reviewers to determine if the generated questions and choices are relevant to the video (Figure 6(a) in the paper).

The second step in the reviewer's verification process is the correctness review. A generated question passes the review if and only if the answers provided by two reviewers are consistent with the GT (Figure 6(b) in the paper).


\begin{figure}[]
    \centering
    \includegraphics[width=\columnwidth]{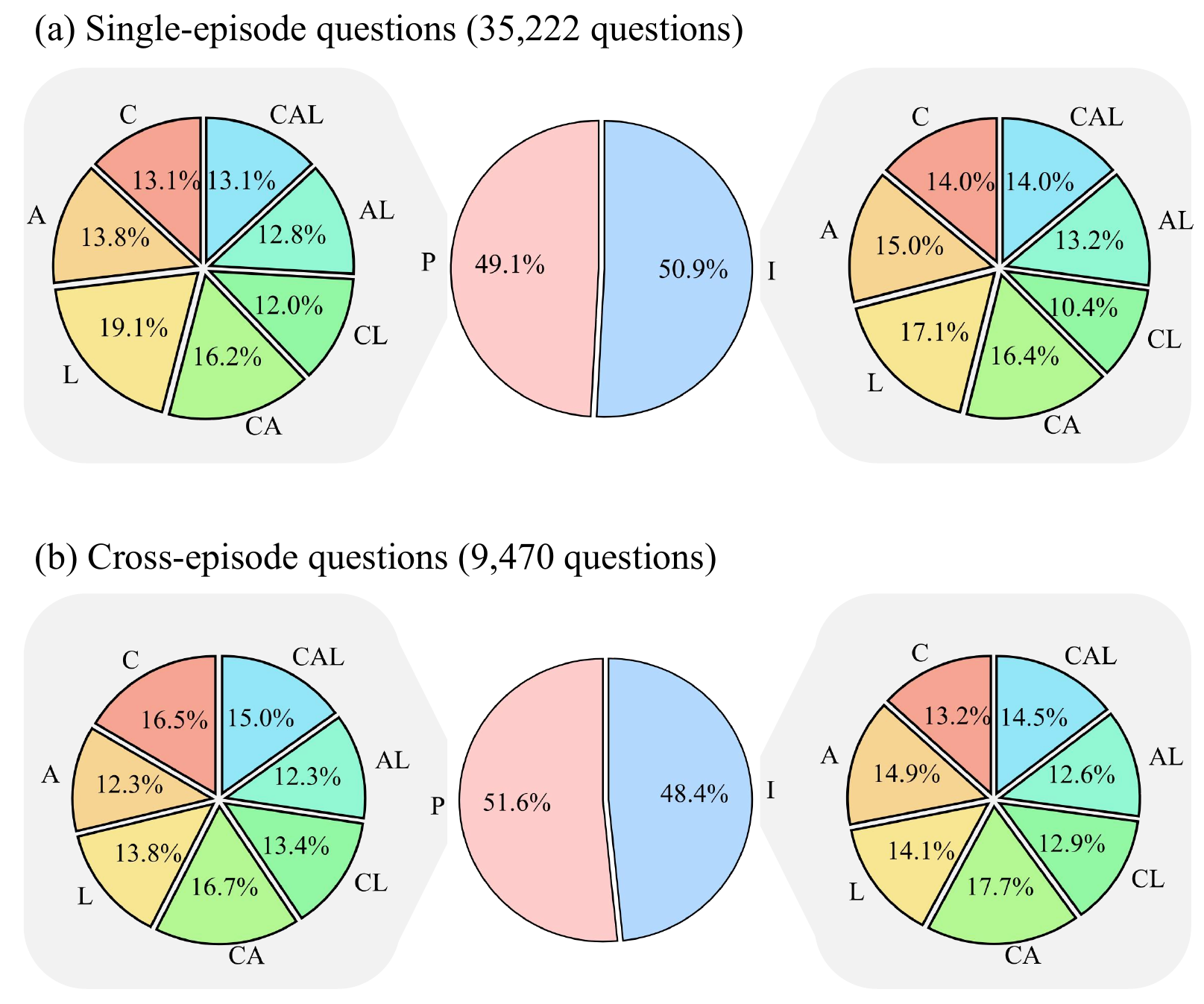}
    \caption{Quantity distributions of single-episode and cross-episode questions of FriendsQA in fine-grained topics}
    \label{fig:scdistribution}
\end{figure}
\section{C. Samples of FriendsQA}

Figure \ref{fig:single_example} and Figure \ref{fig:cross_example} show several single-episode and cross-episode questions in FriendsQA, respectively. For easy verification, we specifically identify each question's fine-grained topic category and difficulty score.


Particularly, for the proposed difficulty measure, we give some samples for different difficulty levels in Figure \ref{fig:factors} to illustrate how time and content factors affect the difficulty measure of the question. For the time factor, the difficulty score and level increase as related time decreases. Similarly, for the content factor, both the difficulty score and level increase as the number of related instances decreases.

\section{D. Quantity Distributions of FriendsQA}

As illustrated in Figure \ref{fig:scdistribution}, the quantity distribution of FriendsQA covers all fine-grained topics and is balanced across them, including single-episode and cross-episode questions.

\begin{figure}[t]
    \centering
    \includegraphics[width=\linewidth]{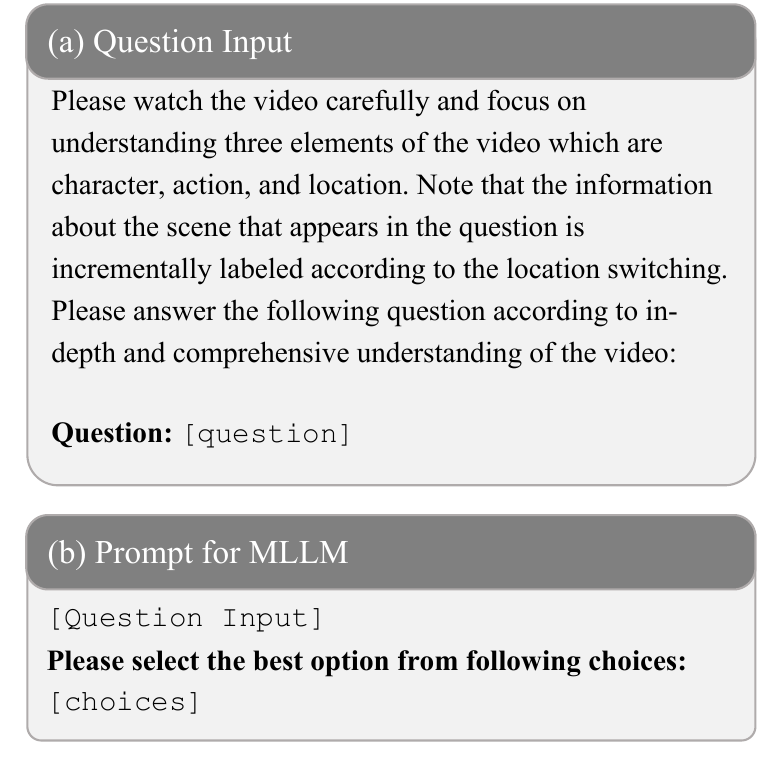}
    \caption{Question input and the prompt for MLLM.}
    \label{fig:qaprompt}
\end{figure}

\section{E. Setting for Evaluation}
The setting for evaluation includes experimental details and the text input of evaluation on FriendsQA. 
\subsection{E.1 Experimental Details}

Following recent VideoQA benchmark \cite{egoschema,mvbench}, we adopt a zero-shot question-answering setting on FriendsQA to test 10 state-of-the-art (SOTA) VideoQA models. To ensure the objectivity of the evaluation, we do not modify the model code and solely utilize the default parameters provided by the official repository for performance evaluation.

\subsection{E.2 Text Input of Evaluation}
This paper evaluates two types of VideoQA models, VLM and MLLM. The similarity between them during evaluation is that they both use video keyframes as the input of visual information, but the difference lies in the input of text information. For VLM, we input question and choices into the question input head and choice input head of the model, respectively. 

For MLLM, we put the question and choices as a whole into the prompt and input them into the model (Figure \ref{fig:qaprompt}). It should be emphasized that in addition to the main question, we also write a description in the question input part to prompt the model to accurately locate the question scene based on the change of locations.


\section{F. Qualitative Experiments}
Figure \ref{fig:qualitative} and Figure \ref{fig:qualitative_cross} present qualitative results on the FriendsQA dataset for 10 SOTA models, encompassing both single-episode and cross-episode questions.

\renewcommand{\arraystretch}{1.4} 
\begin{table*}[t]
\centering
\caption{Question examples of the prompt for question generation.}
\label{tab:qgexample}
\resizebox{\textwidth}{!}{%
\begin{tabular}{cccl}
\toprule
\textbf{Episode} & \textbf{Attribution} & \textbf{Topic} & \textbf{Examples} \\ 
\midrule \midrule
\multirow{14}{*}{\parbox[c][45em][c]{1.5cm}{\centering Single}} & \multirow{7}{*}{\parbox[c][19em][c]{1.5cm}{\centering P}} & C & \begin{tabular}[c]{@{}l@{}}How many important characters from within the TV series appear in Scene 1? \\ In Scene 2, in what order do the characters appear?\end{tabular} \\ \cline{3-4} 
 &  & A & \begin{tabular}[c]{@{}l@{}}What is the main action taking place in Scene 2? \\ What is the order of appearance of actions in Scene 4?\end{tabular} \\ \cline{3-4} 
 &  & L & \begin{tabular}[c]{@{}l@{}}In which location does Scene 5 take place? \\ In Scene 3, where did the event take place?\end{tabular} \\ \cline{3-4} 
 &  & CA & \begin{tabular}[c]{@{}l@{}}What action does Chandler perform while Joey is practicing his lines? \\ Which character takes a puff of a cigarette during the rehearsal scene?\end{tabular} \\ \cline{3-4} 
 &  & CL & \begin{tabular}[c]{@{}l@{}}In Scene 13 which characters are in Central Perk?\\ In which location do Monica and Ross appear together?\end{tabular} \\ \cline{3-4} 
 &  & AL & \begin{tabular}[c]{@{}l@{}}What action is taking place in Scene 3 at the hospital? \\ What sequence of actions takes place in Scene 7 at Central Perk?\end{tabular} \\ \cline{3-4} 
 &  & CAL & \begin{tabular}[c]{@{}l@{}}Who is discussing a topic at Monica's apartment in Scene 12? \\ In Scene 3 at Central Perk, who is sitting, discussing, and then standing up?\end{tabular} \\ \cline{2-4} 
 & \multirow{7}{*}{\parbox[c][19em][c]{1.5cm}{\centering I}} & C & \begin{tabular}[c]{@{}l@{}}Who are the main characters in this video that contribute to the plot?\\ Across the entire episode, how does Rachel's (growth/moods change/opinions)?\end{tabular} \\ \cline{3-4} 
 &  & A & \begin{tabular}[c]{@{}l@{}}In both Scene 1 and Scene 3, what actions appear simultaneously in both scenes? \\ How do the actions related to preparing and consuming meals highlight the friendship throughout the episode?\end{tabular} \\ \cline{3-4} 
 &  & L & \begin{tabular}[c]{@{}l@{}}Throughout the episode, how does Monica's apartment serve as a place for? \\ How do the scenes set at Central Perk contribute to the development of the friendships over the course of the episode?\end{tabular} \\ \cline{3-4} 
 &  & CA & \begin{tabular}[c]{@{}l@{}}In Scene 2, after Joey greets everyone, what action do the other characters perform?\\ How do Ross's interactions and actions with Monica and Phoebe in various scenes depict her role within the friend group?\end{tabular} \\ \cline{3-4} 
 &  & CL & \begin{tabular}[c]{@{}l@{}}Which characters appear at Chandler's apartment in Scene 4 and also appear at Central Perk in Scene 14? \\ Which characters appear in both Monica and Rachel's apartment in Scene 12 and Central Perk in Scene 5?\end{tabular} \\ \cline{3-4} 
 &  & AL & \begin{tabular}[c]{@{}l@{}}What actions happen in locations of Scene 12 and Scene 15?\\ Across multiple scenes at Central Perk, what actions reflect the reliance on this location for emotional support?\end{tabular} \\ \cline{3-4} 
 &  & CAL & \begin{tabular}[c]{@{}l@{}}In Scene 6 at the street, who gives \$1000 and a football phone, and what action do they perform next?\\ How do the interactions at Central Perk and Monica's apartment across various scenes reveal the core dynamics of the friends group?\end{tabular} \\ \hline
\multirow{14}{*}{\parbox[c][45em][c]{1.5cm}{\centering Cross}} & \multirow{7}{*}{\parbox[c][19em][c]{1.5cm}{\centering P}} & C & \begin{tabular}[c]{@{}l@{}}Which two characters are shown to be brother and sister? \\ In these episodes, which two characters fall in love with each other?\end{tabular} \\ \cline{3-4} 
 &  & A & \begin{tabular}[c]{@{}l@{}}What recurring action is performed whenever someone enters their apartment? \\ What action is considered a symbolic gesture of leaving her old life of depending on her family?\end{tabular} \\ \cline{3-4} 
 &  & L & \begin{tabular}[c]{@{}l@{}}How many locations appears in these episodes? \\ Which location appears most in these episodes?\end{tabular} \\ \cline{3-4} 
 &  & CA & \begin{tabular}[c]{@{}l@{}}What action does Ross repeatedly perform as he reminisces about his relationship with Carol?\\ Which character is the most likely to use humor as a defense mechanism?\end{tabular} \\ \cline{3-4} 
 &  & CL & \begin{tabular}[c]{@{}l@{}}Which character is shown his/her work in Central Perk throughout all episodes?\\ In which locations does Ross reveal his past with Carol?\end{tabular} \\ \cline{3-4} 
 &  & AL & \begin{tabular}[c]{@{}l@{}}What location is frequently used for watching TV? \\ What action is repeatedly performed in both Monica's apartment and Central Perk?\end{tabular} \\ \cline{3-4} 
 &  & CAL & \begin{tabular}[c]{@{}l@{}}Who is shown working at Central Perk after giving up their credit cards? \\ At which locations does Chandler smoke a cigarette?\end{tabular} \\ \cline{2-4} 
 & \multirow{7}{*}{\parbox[c][19em][c]{1.5cm}{\centering I}} & C & \begin{tabular}[c]{@{}l@{}}Which two characters are shown to be twins or brother and sister? \\ Who appears to be the most pragmatic and realistic in their approach to life?\end{tabular} \\ \cline{3-4} 
 &  & A & \begin{tabular}[c]{@{}l@{}}How do the reactions to receiving unexpected gifts reveal the beliefs?\\ How does the recurring action of 'hugging' demonstrate the importance of physical touch and intimacy in their friendships?\end{tabular} \\ \cline{3-4} 
 &  & L & \begin{tabular}[c]{@{}l@{}}How do the different apartments featured in the episodes reflect the personalities and lifestyles?\\ How does the setting of Central Perk contribute to the overall tone and atmosphere of the episodes?\end{tabular} \\ \cline{3-4} 
 &  & CA & \begin{tabular}[c]{@{}l@{}}Based on their interactions, which character seems to be the most empathetic and understanding of others' feelings?\\ Why does Monica keep her relationship with Alan a secret from her friends for so long?\end{tabular} \\ \cline{3-4} 
 &  & CL & \begin{tabular}[c]{@{}l@{}}How does the contrast between Monica and Rachel's apartment and Ross's apartment reflect their differing emotional states? \\ How does the setting of Monica and Rachel's apartment reflect the evolving dynamics of their friendship?\end{tabular} \\ \cline{3-4} 
 &  & AL & \begin{tabular}[c]{@{}l@{}}How does the location of the office contribute to the portrayal of professional life and personal struggles?\\ At which location frequently discuss romantic relationships and drinking coffee?\end{tabular} \\ \cline{3-4} 
 &  & CAL & \begin{tabular}[c]{@{}l@{}}Which character works in a restaurant but dreams of having their own restaurant?\\ Why is Monica's apartment the primary location for the group's emotional discussions and support?\end{tabular} \\ \bottomrule
\end{tabular}%
}
\end{table*}

\begin{figure*}[t]
    \centering
    \includegraphics[width=\linewidth]{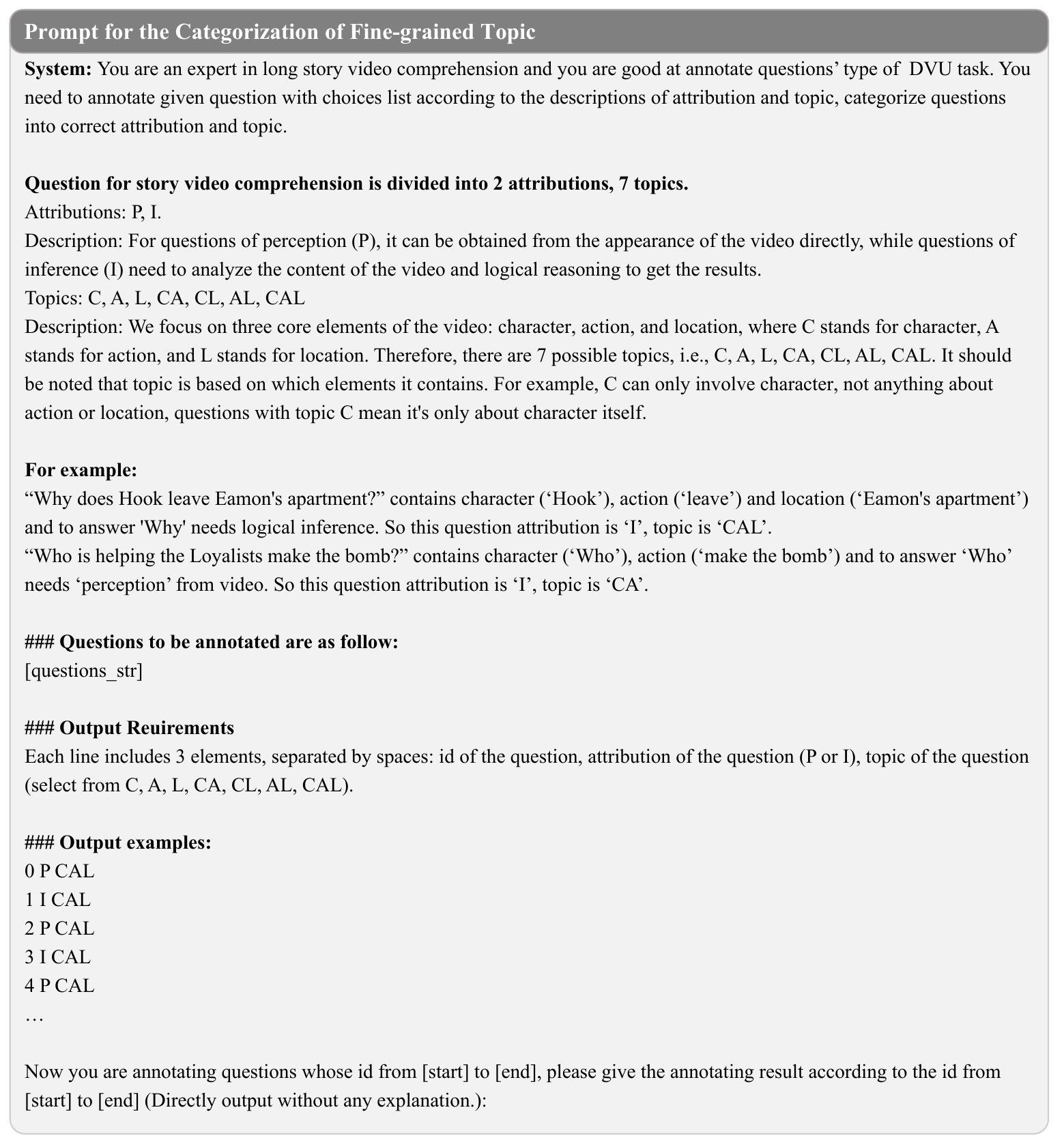}
    \caption{The prompt template for assessing fine-grained topics of different DVU datasets.}
    \label{fig:top_cat}
\end{figure*}

\begin{figure*}[t]
    \centering
    \includegraphics[width=\linewidth]{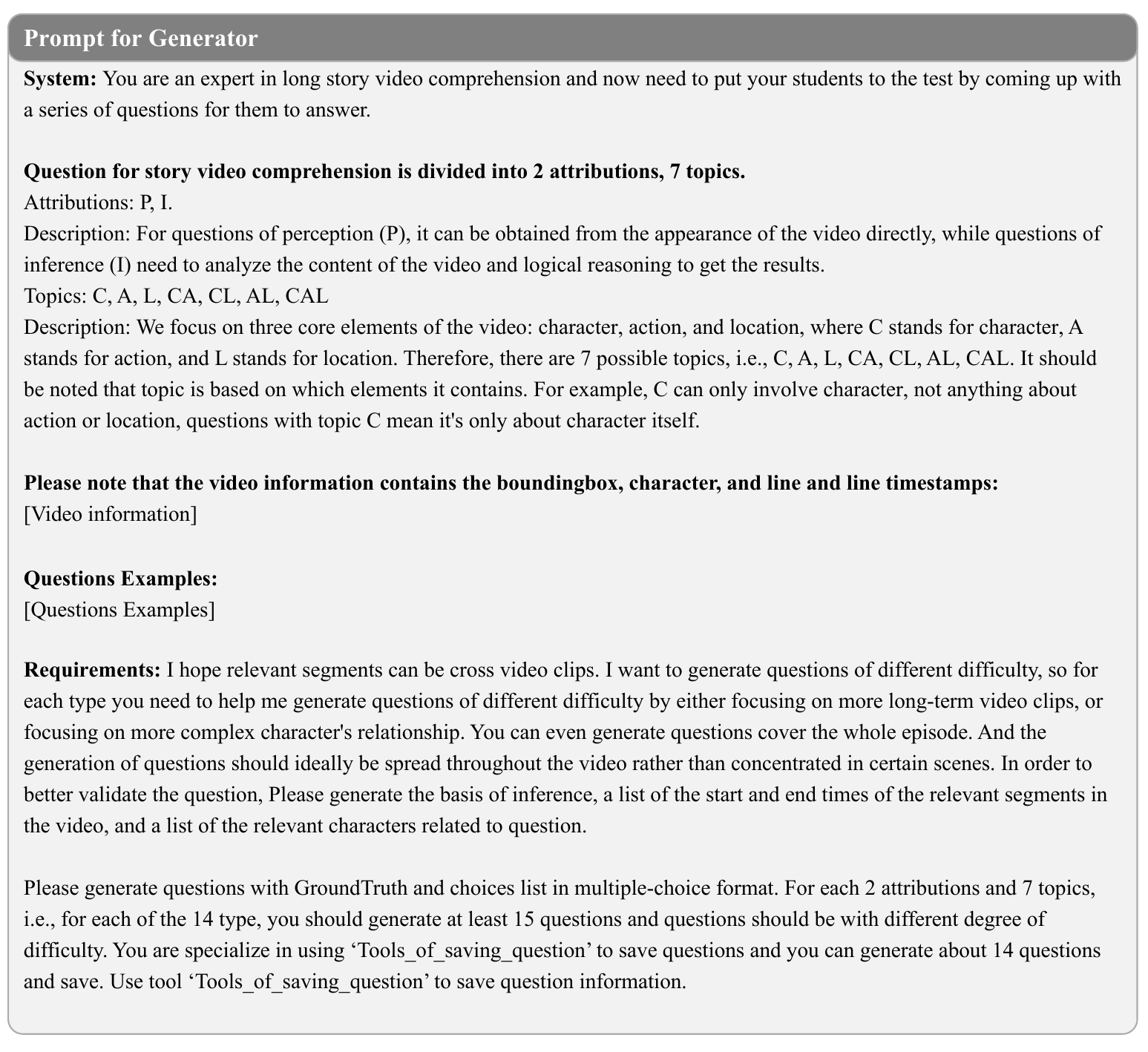}
    \caption{The prompt template for generator.}
    \label{fig:generator}
\end{figure*}
\begin{figure*}[t]
    \centering
    \includegraphics[width=\linewidth]{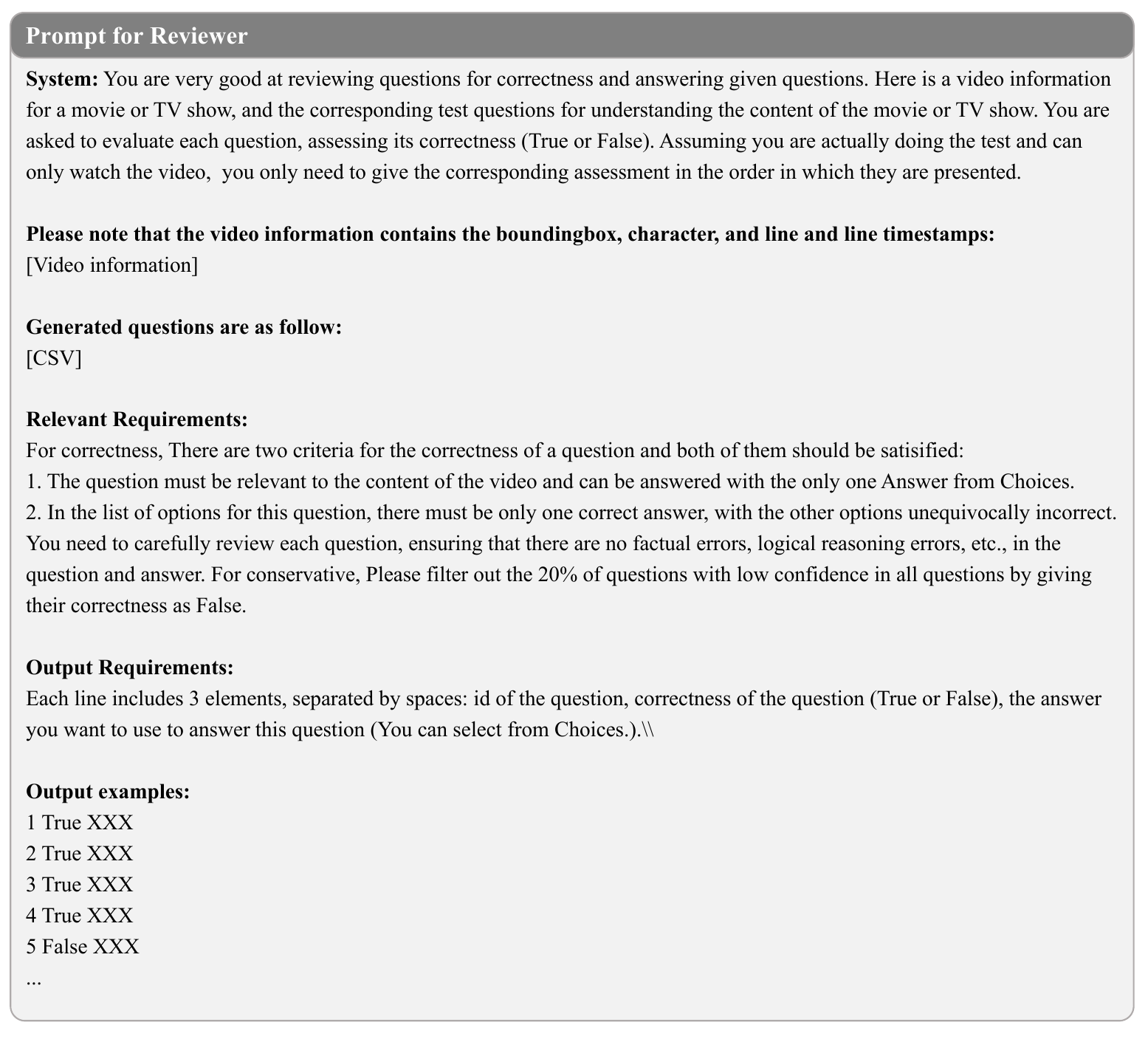}
    \caption{The prompt template for reviewer.}
    \label{fig:reviewer_prompt}
\end{figure*}
\begin{figure*}[t]
    \centering
    \includegraphics[width=\textwidth]{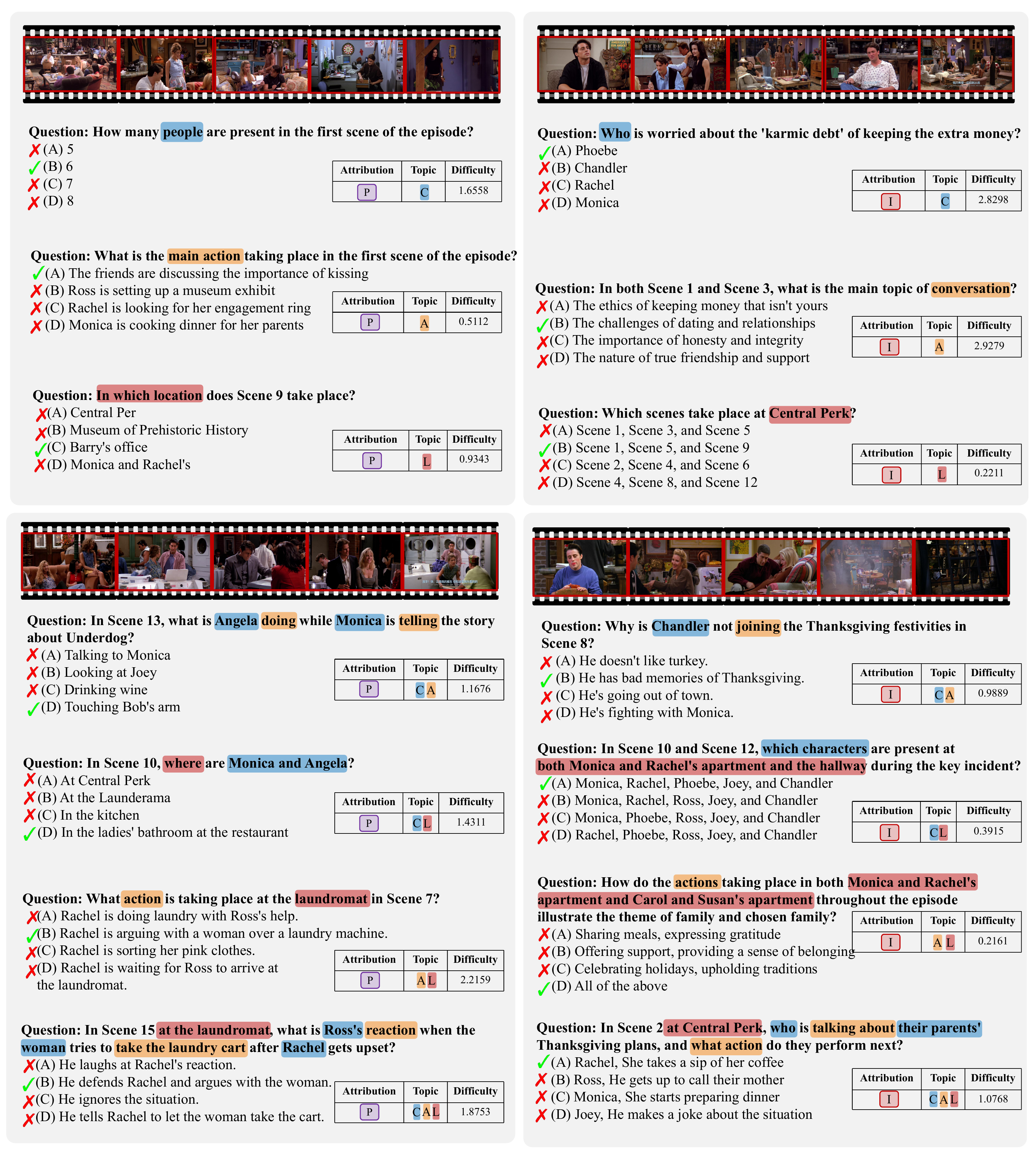}
    \caption{Examples of the fine-grained topic questions (single-episode questions)}
    \label{fig:single_example}
\end{figure*}
\begin{figure*}[t]
    \centering
    \includegraphics[width=\textwidth]{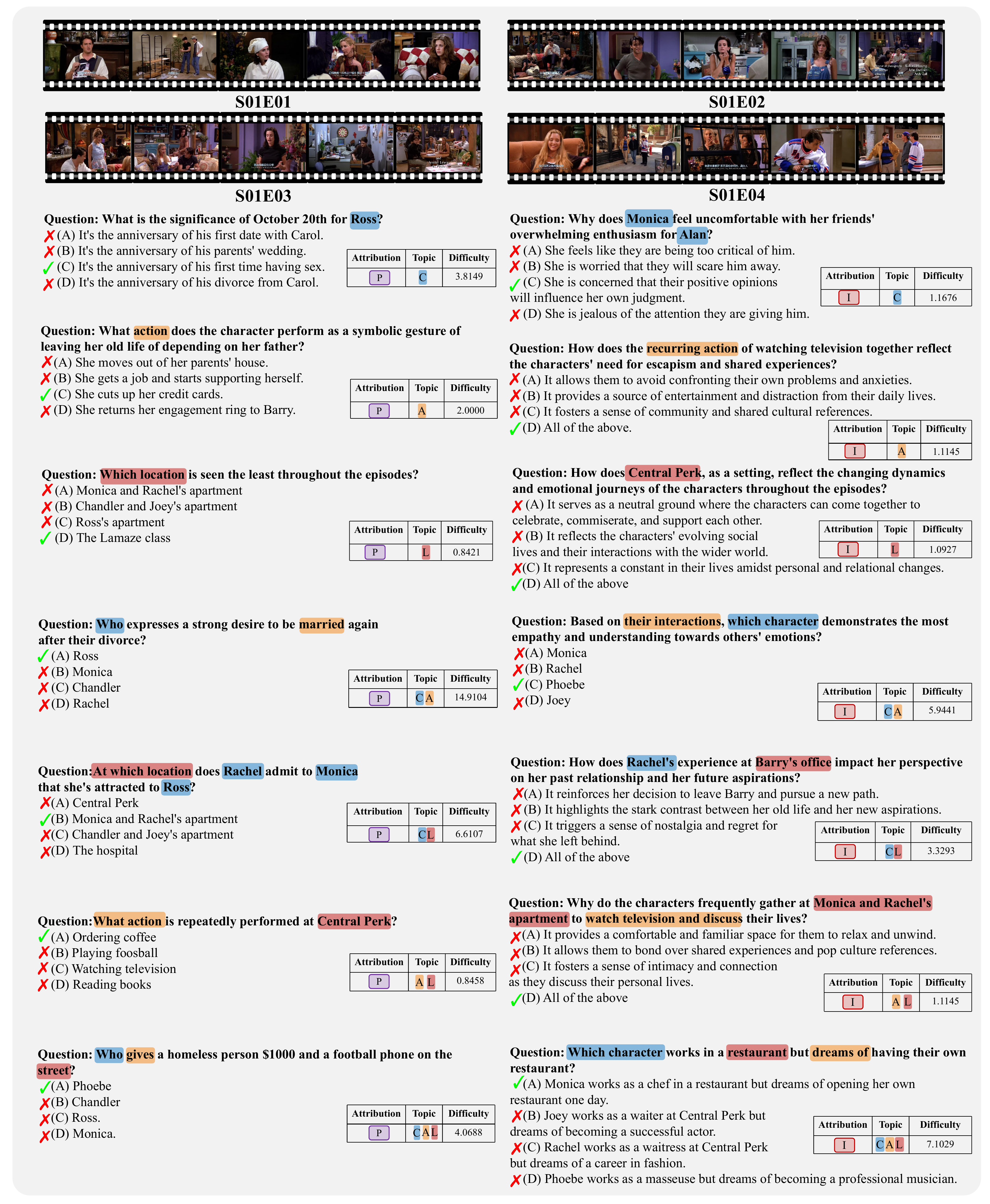}
    \caption{Examples of the fine-grained topic questions (cross-episode questions)}
    \label{fig:cross_example}
\end{figure*}
\begin{figure*}[t]
    \centering
    \includegraphics[width=\textwidth]{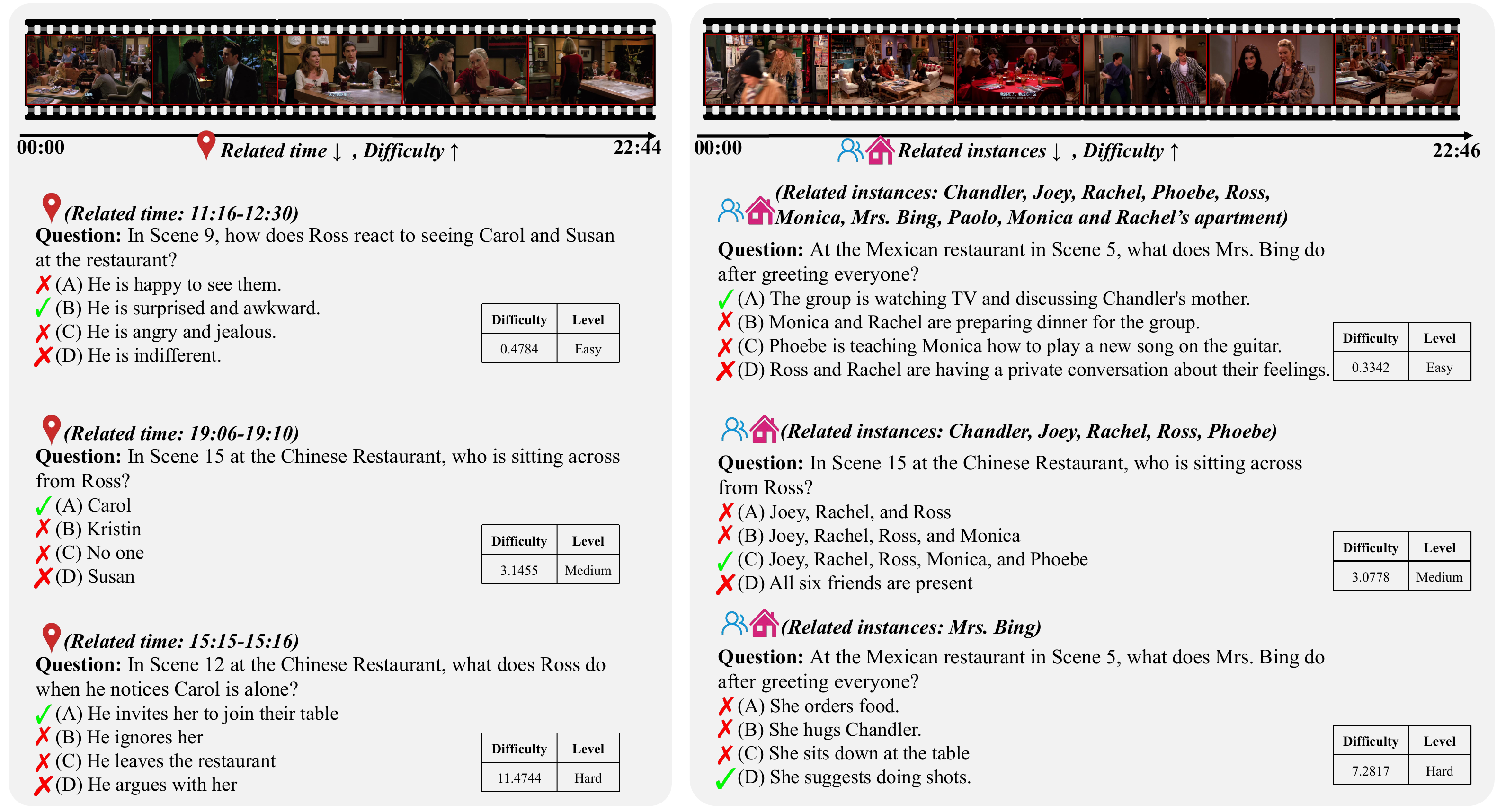}
    \caption{Samples of FriendsQA in different difficulty levels}
    \label{fig:factors}
\end{figure*}

\begin{figure*}[t]
    \centering
    \includegraphics[width=0.9\textwidth]{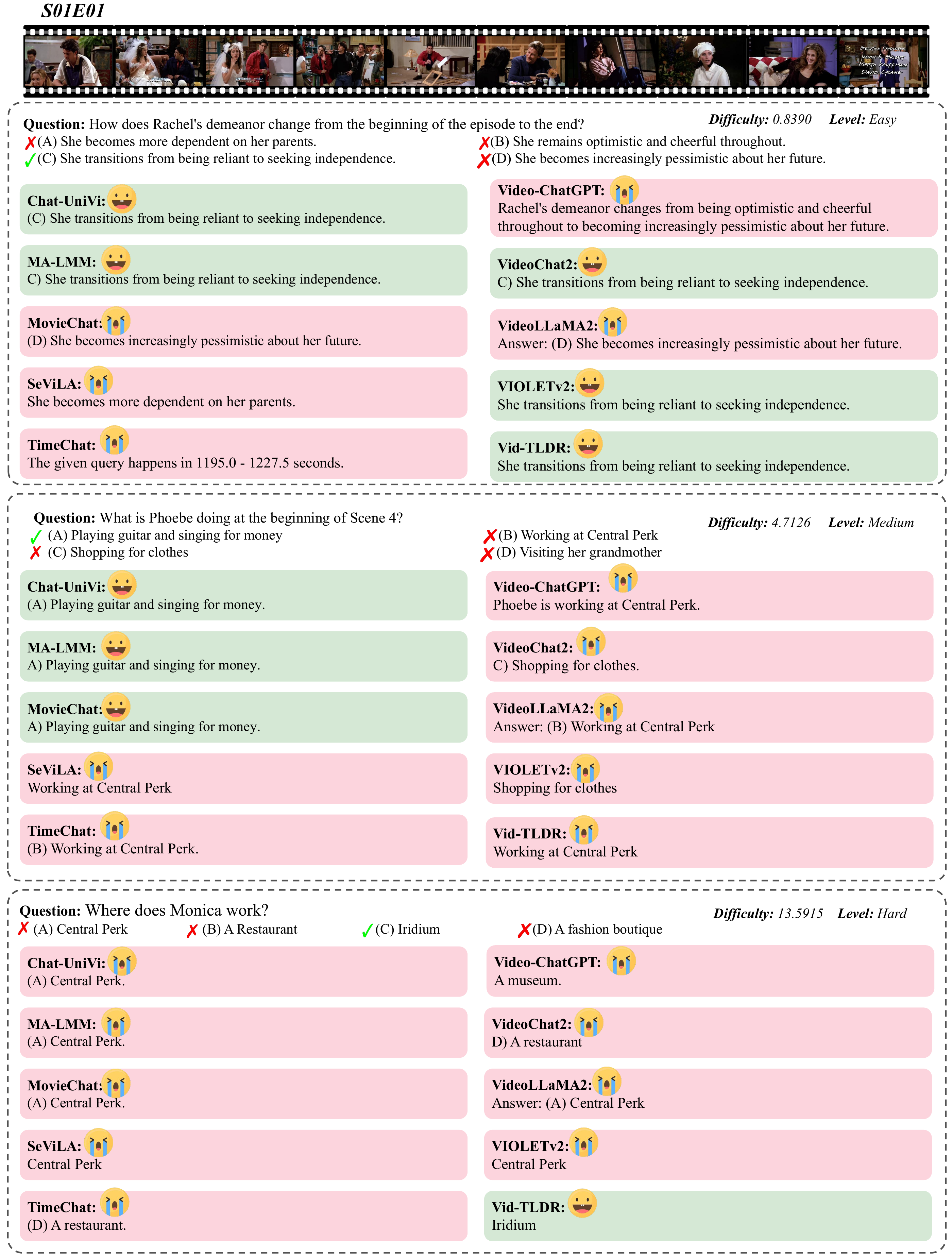}
    \caption{Qualitative results across 10 SOTA models on single-episode questions.}
    \label{fig:qualitative}
\end{figure*}

\begin{figure*}[t]
    \centering
    \includegraphics[width=0.9\textwidth]{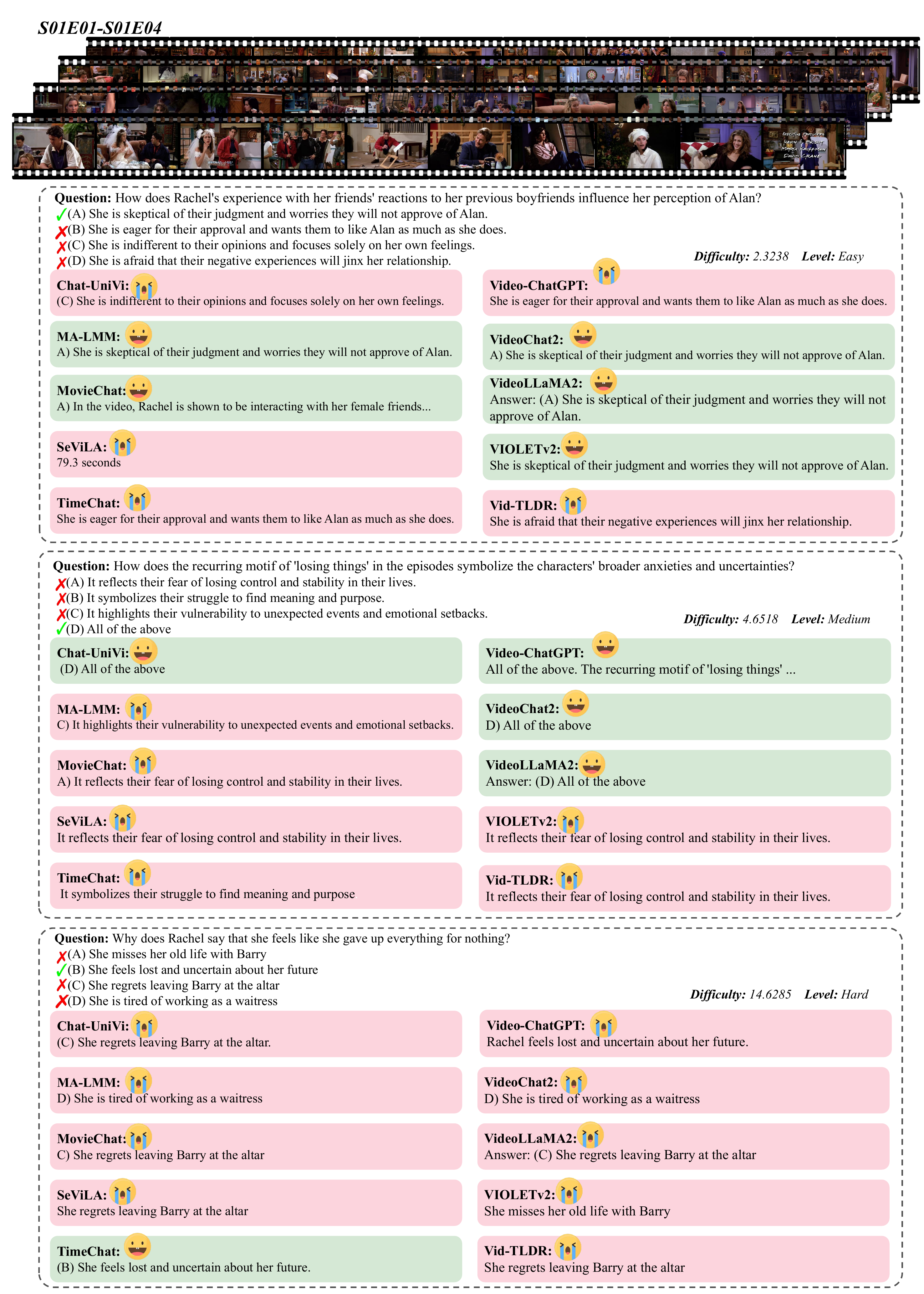}
    \caption{Qualitative results across 10 SOTA models on cross-episode questions.}
    \label{fig:qualitative_cross}
\end{figure*}

\end{document}